\documentclass[10pt,twocolumn,letterpaper]{article}

\usepackage{cvpr}              

\usepackage{rotating}
\usepackage{multirow}

\usepackage[table]{xcolor} 
\usepackage{booktabs}      
\newcommand{\first}[1]{\cellcolor[HTML]{63C07A}#1}   
\newcommand{\second}[1]{\cellcolor[HTML]{BDDC8E}#1}  
\newcommand{\third}[1]{\cellcolor[HTML]{EBE7A0}#1}            
\usepackage[most]{tcolorbox} 
\usepackage[dvipsnames, table]{xcolor}
\definecolor{firstColor}{HTML}{63C07A}
\definecolor{secondColor}{HTML}{BDDC8E}
\definecolor{thirdColor}{HTML}{EBE7A0}
\newcommand{\First}[1]{\colorbox{firstColor}{#1}}
\newcommand{\Second}[1]{\colorbox{secondColor}{#1}}
\newcommand{\Third}[1]{\colorbox{thirdColor}{#1}}

\usepackage{color}
\definecolor{red}{rgb}{1.00,0.00,0.00}
\definecolor{blue}{rgb}{0.00,0.00,1.00}
\definecolor{green}{rgb}{0.04,0.40,0.14}
\definecolor{yellow}{rgb}{0.5,0.5,0.0}
\definecolor{purple}{rgb}{0.5,0.0,0.5}
\newcommand{\cred}[1] {\textcolor{red}{#1}}

\definecolor{cvprblue}{rgb}{0.21,0.49,0.74}
\usepackage[pagebackref,breaklinks,colorlinks,allcolors=cvprblue]{hyperref}

\def\paperID{*****} 
\def\confName{CVPR}
\def\confYear{2026}

\title{Camo-M3FD: A New Benchmark Dataset for Cross-Spectral Camouflaged Pedestrian Detection
}

\author{Henry O. Velesaca$^{1,3}$, Andrea Mero$^{1,4}$, Guillermo A. Castillo$^{1}$, Angel D. Sappa$^{1,2}$ \vspace{2mm}\\
$^{1}$ESPOL Polytechnic University, Campus Gustavo Galindo, 090902, Guayaquil, Ecuador  \vspace{2mm}\\
$^2$Computer Vision Center, Universitat Aut\`onoma de Barcelona, 08193-Bellaterra, Barcelona, Spain \vspace{2mm}\\
$^3$Software Engineering Department, Research Center for Information and Communication \\ Technologies (CITIC-UGR), University of Granada, 18071, Granada, Spain \vspace{2mm}\\
$^4$Università della Svizzera Italiana, Via Giuseppe Buffi 13, 6900, Lugano, Switzerland \vspace{2mm}\\
{\tt\small \{hvelesac,anmero,guancast\}@espol.edu.ec, sappa@ieee.org}
}

\begin{document}
\maketitle

\begin{abstract}
Pedestrian detection is fundamental to autonomous driving, robotics, and surveillance. Despite progress in deep learning, reliable identification remains challenging due to occlusions, cluttered backgrounds, and degraded visibility. While multispectral detection—combining visible and thermal sensors—mitigates poor visibility, the challenge of camouflaged pedestrians remains largely unexplored. Existing Camouflaged Object Detection (COD) benchmarks focus on biological species, leaving a gap in safety-critical human detection where targets blend into their surroundings. To address this, we introduce Camo-M3FD (derived from the M3FD dataset), a novel benchmark for cross-spectral camouflaged pedestrian detection, consisting of registered visible-thermal image pairs. The dataset is curated using quantitative metrics to ensure high foreground-background similarity. We provide high-quality pixel-level masks and establish a standardized evaluation framework using state-of-the-art COD models. Our results demonstrate that while thermal signals provide indispensable localization cues, multispectral fusion is essential for refining structural details. Camo-M3FD serves as a foundational resource for developing robust and safety-critical detection systems. The dataset is available on GitHub: \url{https://cod-espol.github.io/Camo-M3FD/}.
\end{abstract}

\section{Introduction}
\label{sec:intro}
Pedestrian detection has been fundamental to the development of autonomous driving, robotics, and modern surveillance. However, despite the remarkable progress of deep learning–based detectors, reliable pedestrian detection in real-world scenarios remains challenging due to occlusions, cluttered backgrounds, diverse human poses, scale variation, and—critically—visibility degradation caused by nighttime conditions and adverse weather.\cite{ghari2024pedestrian, chen2023occlusion, tumas2020pedestrian}

To mitigate poor visibility, multispectral sensing has become a common strategy, combining visible-spectrum cameras with thermal sensors to acquire aligned image pairs \cite{teutsch2021computer}. Works such as \cite{hwang2015multispectral} have shown that thermal information provides complementary cues for pedestrians when visual appearance is weak (e.g., at night, at far range, under partial occlusion, or across challenging weather conditions).

Beyond “difficult visibility” in general, camouflage introduces an even harder challenge, since targets are visually designed (or naturally appear) to blend into their surroundings. COD explicitly addresses this regime by focusing on objects that merge with background appearance and therefore require more sophisticated strategies than conventional detection paradigms. Surveys such as \cite{xiao2024survey, zhong2024survey} highlight that widely used COD datasets predominantly feature animals, leaving only a small fraction of samples containing people. Moreover, part of those samples involve military-grade camouflage, and in \cite{liu2023camouflaged} it can be observed that the ground-truth annotations in these military-camouflage datasets often cover not only the person but also weapons. This further reduces the fraction of images that are truly relevant for human-centered operational settings (e.g., surveillance, security, or search and rescue). As a result, camouflaged pedestrian detection remains under-explored—a critical gap given that pedestrians are safety-critical targets and failures can have severe real-world consequences.

This work introduces Camo-M3FD, a new benchmark dataset for cross-spectral camouflaged pedestrian detection developed from registered visible–thermal image pairs. The dataset is curated using quantitative camouflage metrics that capture foreground–background similarity in color, appearance, and boundary consistency to specifically retain pedestrian instances with strong environmental blending. In addition, high-quality pixel-level semantic ground-truth masks are provided. To enable standardized and comprehensive evaluation, a suite of widely recognized COD metrics is adopted, encompassing structural similarity, boundary accuracy, and global alignment. Furthermore, performance is reported using various adaptive and threshold-agnostic measures derived from precision–recall curves to ensure a representative and robust assessment of model capabilities.

The manuscript is organized as follows. Section~\ref{sec:back} introduces related work, recent SOTA COD techniques, and methods that address the problem of the COD approach. Section~\ref{sec:prop} presents the methodology to construct the dataset. Then, Section~\ref{sec:exp} shows the experimental results using different COD techniques. Finally, discussion and conclusions are given in Section~\ref{sec:disc} and Section~\ref{sec:conclu} respectively.

\section{Related Works}
\label{sec:back}
This section situates the Camo-M3FD benchmark within the broader evolution of pedestrian detection and camouflaged object analysis. It reviews classical human detection methodologies, from handcrafted descriptors to early multispectral integrations, alongside contemporary state-of-the-art COD techniques. By analyzing how edge modeling and cross-spectral fusion address environmental blending, the theoretical foundation for high-precision, cross-spectral camouflaged pedestrian detection is established.

\subsection{Classical Pedestrian Detection Techniques}
Early pedestrian detection approaches were largely organized around a multi-scale sliding-window paradigm, where window-wise descriptors are extracted, a binary classifier is applied, a dense search over position/scale is performed, and results are finalized with non-maximum suppression (NMS) (e.g., \cite{geronimo2006pedestrian}, \cite{geronimo2007computer}); moreover, \cite{ dollar2011pedestrian} argues for its suitability in low- to medium-resolution scenarios where segmentation- or keypoint-based methods tend to fail.

A major shift occurred with gradient-based descriptors, where the Histogram of Oriented Gradients (HOG) combined with linear SVMs became a strong baseline for pedestrian detection due to its sensitivity to the structure of the human silhouette; additionally, texture cues such as Local Binary Patterns (LBP) were fused with HOG to improve robustness to illumination and pose changes \cite{brunetti2018computer, dalal2005histograms}.

To better handle articulation and occlusion, part-based formulations emerged, ranging from supervised part detectors to Deformable Part Models (DPM) that rely on a multi-scale HOG pyramid; learning is formulated via latent SVM supported by hard-negative mining; moreover, variants incorporating dimensionality reduction through PCA and multi-scale refinements report performance gains on benchmarks such as Inria \cite{dalal2005histograms} and Caltech \cite{dollar2011pedestrian, dollar2009pedestrian}. However, these approaches typically incur higher computational cost and are less suitable for real-time deployment without significant approximations \cite{brunetti2018computer, felzenszwalb2008discriminatively}.

The work in \cite{hwang2015multispectral} is relevant to our setting because it couples a multispectral benchmark with a strong classical baseline. The authors introduced a color--thermal pedestrian dataset captured with beam splitter-based hardware to physically align the two image domains, and proposed multispectral extensions of ACF by incorporating thermal-derived channels alongside conventional color/gradient channels. Their analysis shows that thermal cues can be particularly useful in challenging cases, such as long-range pedestrians and partial occlusions, establishing a practical reference point for cross-spectral pedestrian detection.

\subsection{Camouflaged Object Detection Techniques}
This section reviews contemporary state-of-the-art (SoTA) techniques in COD, focusing on diverse strategies designed to overcome the challenges of low-contrast boundaries and environmental blending.

The integration of explicit edge modeling and sophisticated feature fusion mechanisms characterizes recent advancements. BASNet \cite{qin2019basnet} utilizes a predict-and-refine framework with a hybrid loss; however, it relies on structural similarity (SSIM) for implicit edge enhancement rather than explicit supervision. Conversely, SINet \cite{fan2021concealed} adopts a biologically inspired search-and-identification pipeline that leverages edge cues to localize targets. Similarly, EAMNet \cite{sun2023edge} and BGNet \cite{chen2022boundary} incorporate dedicated parallel branches or guidance modules to model object contours directly, while DGNet \cite{ji2023deep} captures edges via gradient flow to detect subtle contrast variations in low-texture regions. For global-local feature integration, CTF-Net \cite{zhang2025effective} combines CNN-based local features with Transformer-based global context to improve boundary precision. In contrast, C$^{2}$F-Net \cite{chen2022camouflaged} employs cross-level context fusion to reinforce structural coherence.

The challenge of cross-spectral and domain-specific detection has led to more specialized architectures. AVNet \cite{velesaca2026iguana} introduces a cross-spectral attention-vision model specifically designed for ecological conservation, effectively fusing visible and infrared information to identify camouflaged targets in complex natural habitats. In parallel, PCNet \cite{yang2024plantcamo} targets plant-specific camouflage through multi-scale refinement, addressing irregular edges characteristic of vegetation. Finally, iterative approaches such as HitNet \cite{hu2023high} and uncertainty-aware models such as OCENet \cite{liu2022modeling} provide dynamic supervision of high-uncertainty regions, thereby indirectly improving boundary clarity. Collectively, these methods represent a paradigm shift toward balancing explicit geometric guidance with implicit contextual refinement across diverse modality scenarios. 

\begin{figure*}[!h]
\setlength\tabcolsep{0.75pt}
\centering
\scalebox{1.0}{
\begin{tabular}{ccccc}

\includegraphics[width=.17\textwidth, height=2.0cm]{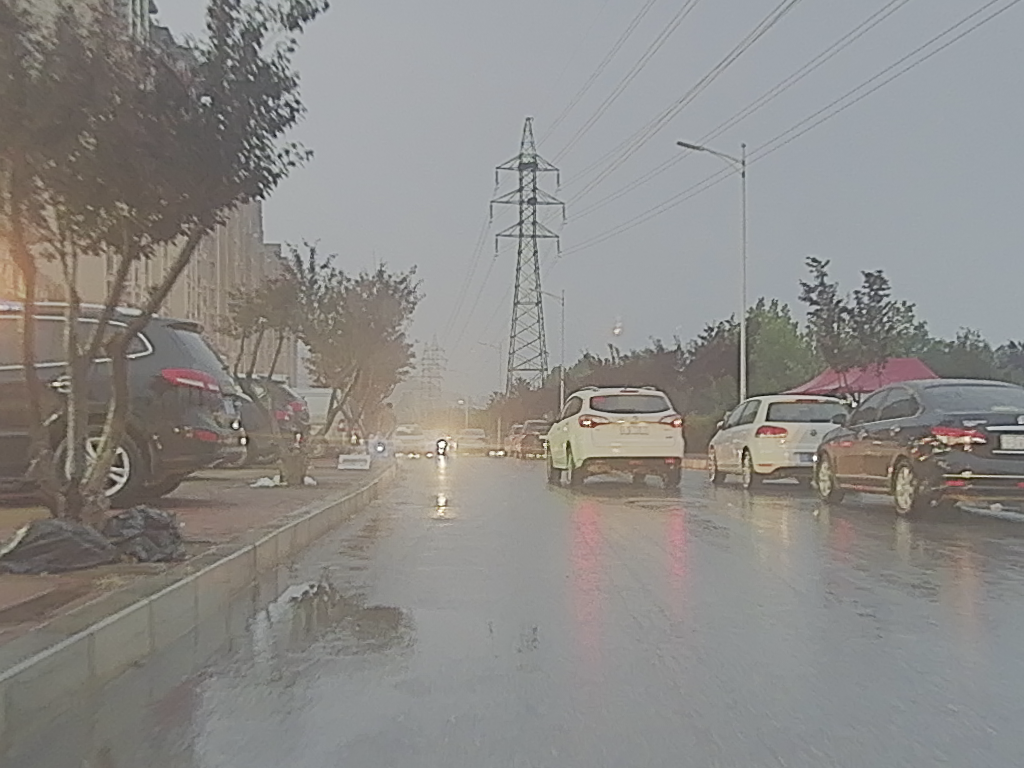} &
\includegraphics[width=.17\textwidth, height=2.0cm]{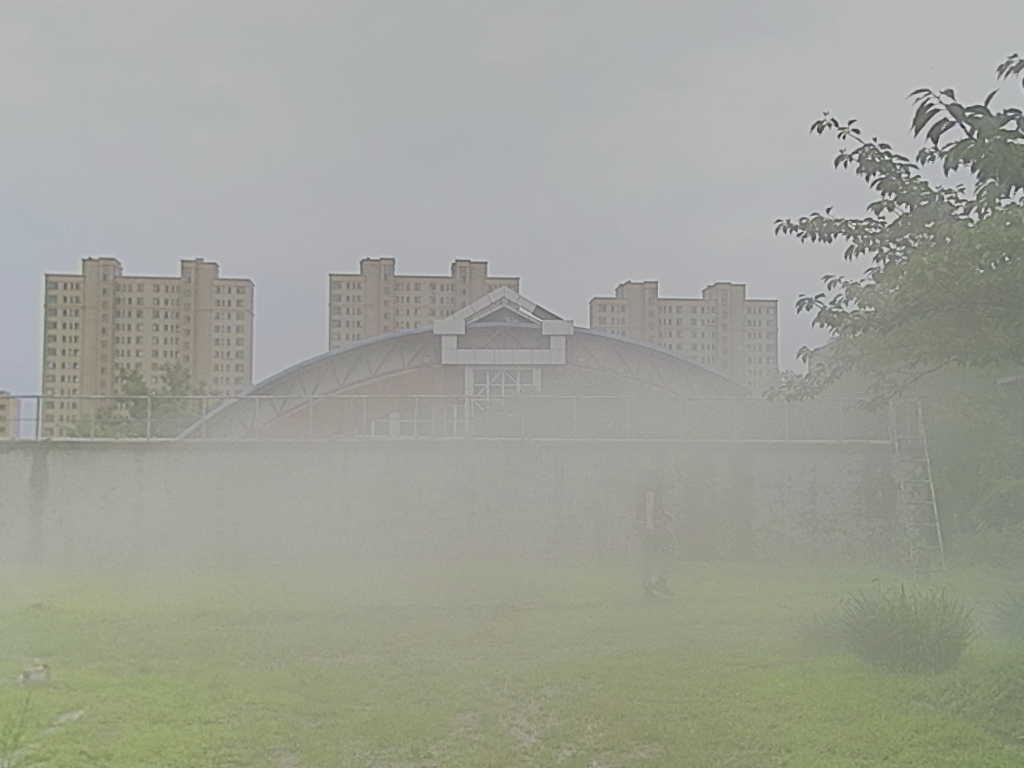} &
\includegraphics[width=.17\textwidth, height=2.0cm]{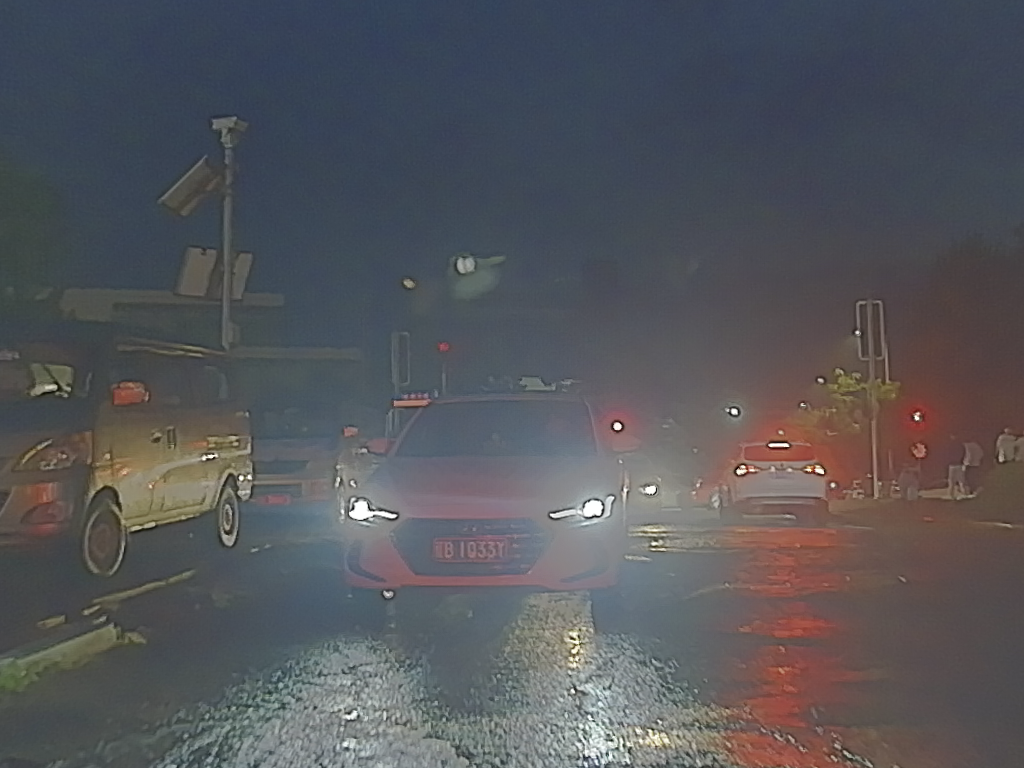} & 
\includegraphics[width=.17\textwidth, height=2.0cm]{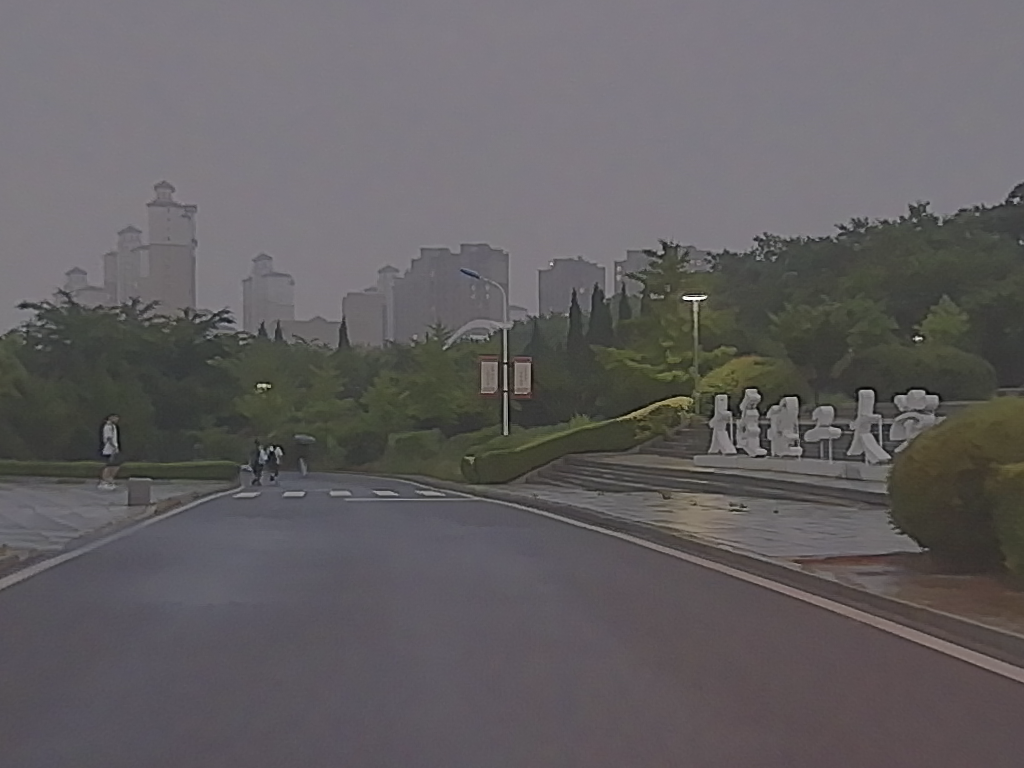} &
\includegraphics[width=.17\textwidth, height=2.0cm]{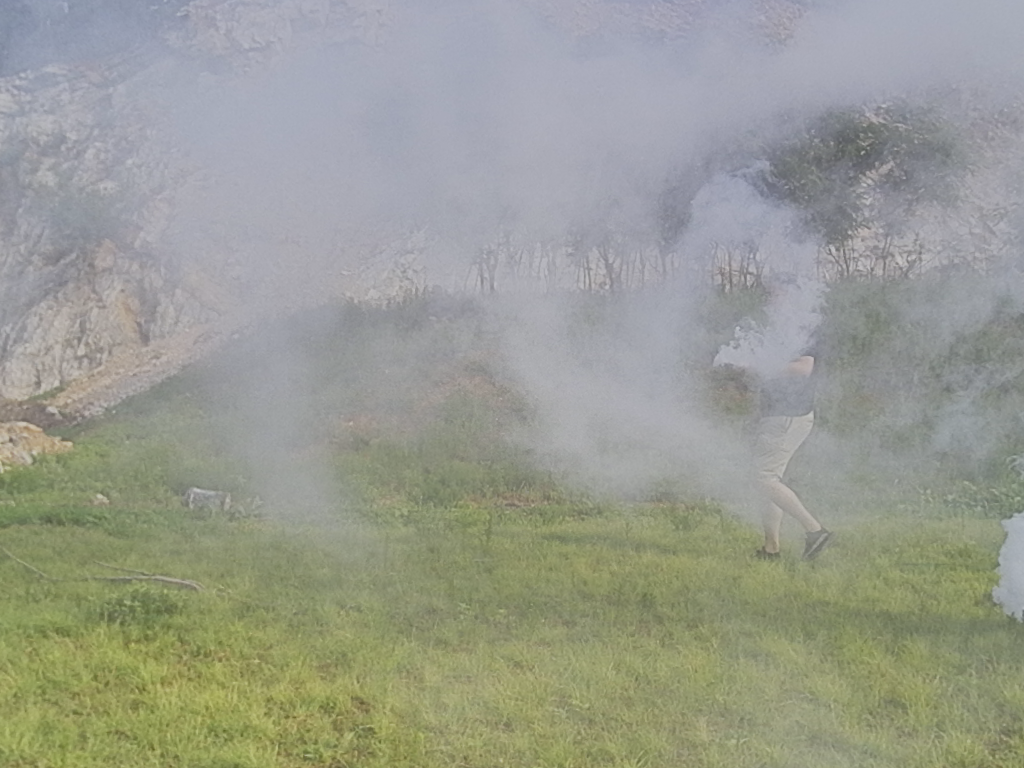} \\

\includegraphics[width=.17\textwidth, height=2.0cm]{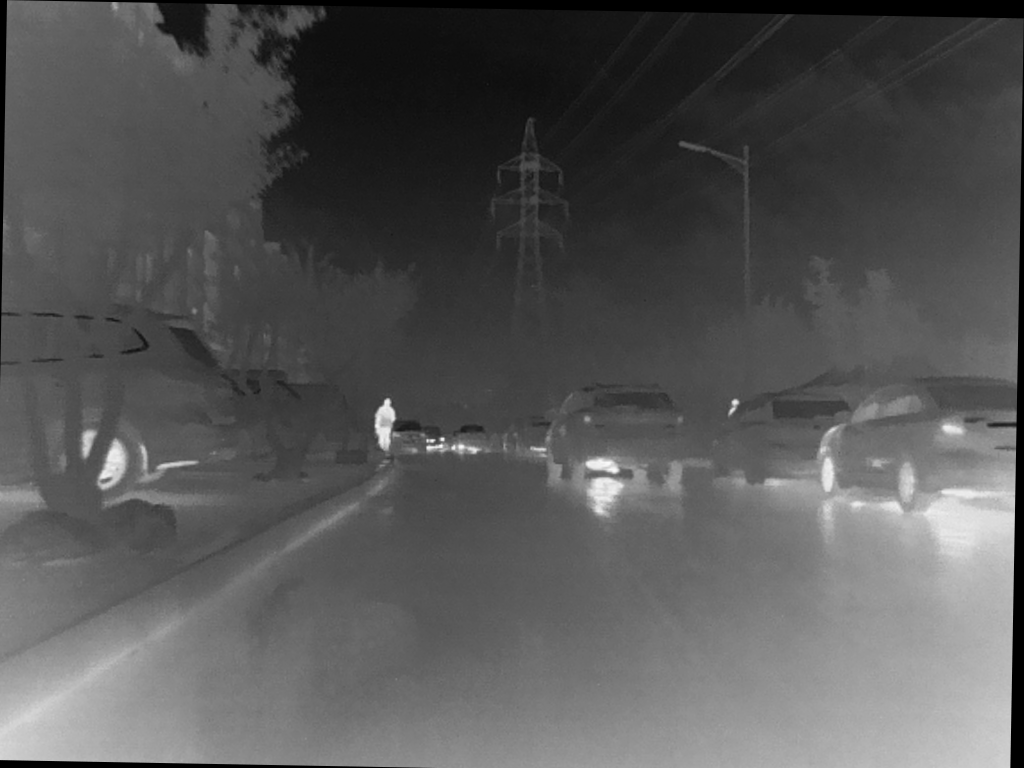} &
\includegraphics[width=.17\textwidth, height=2.0cm]{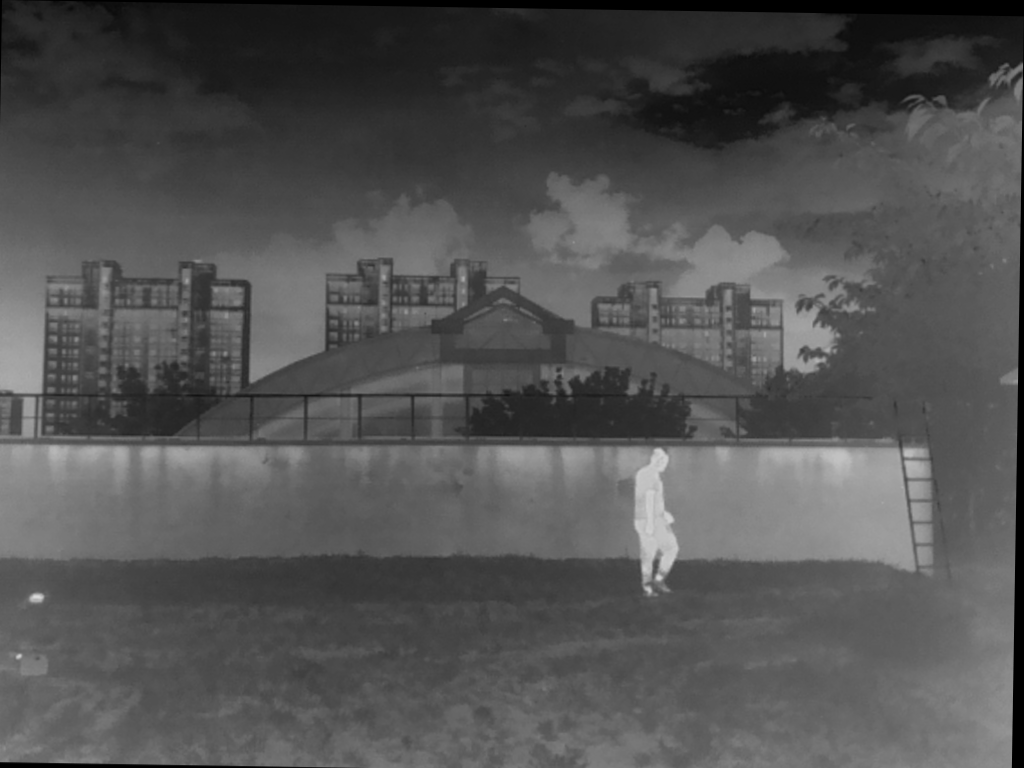} &
\includegraphics[width=.17\textwidth, height=2.0cm]{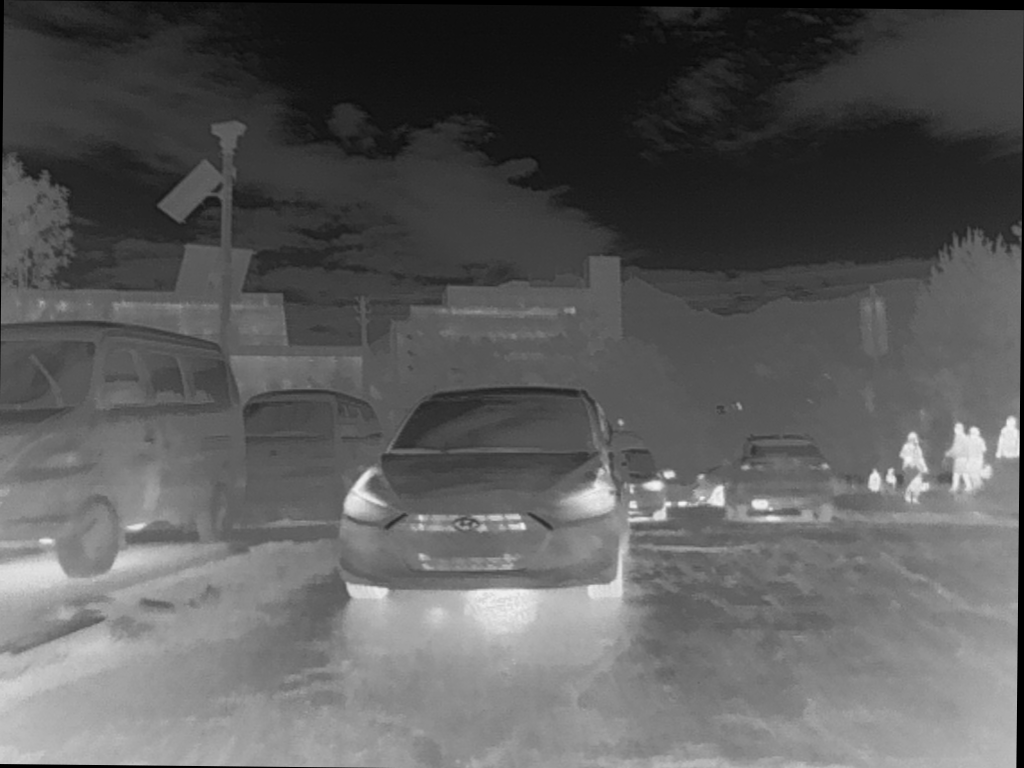} & 
\includegraphics[width=.17\textwidth, height=2.0cm]{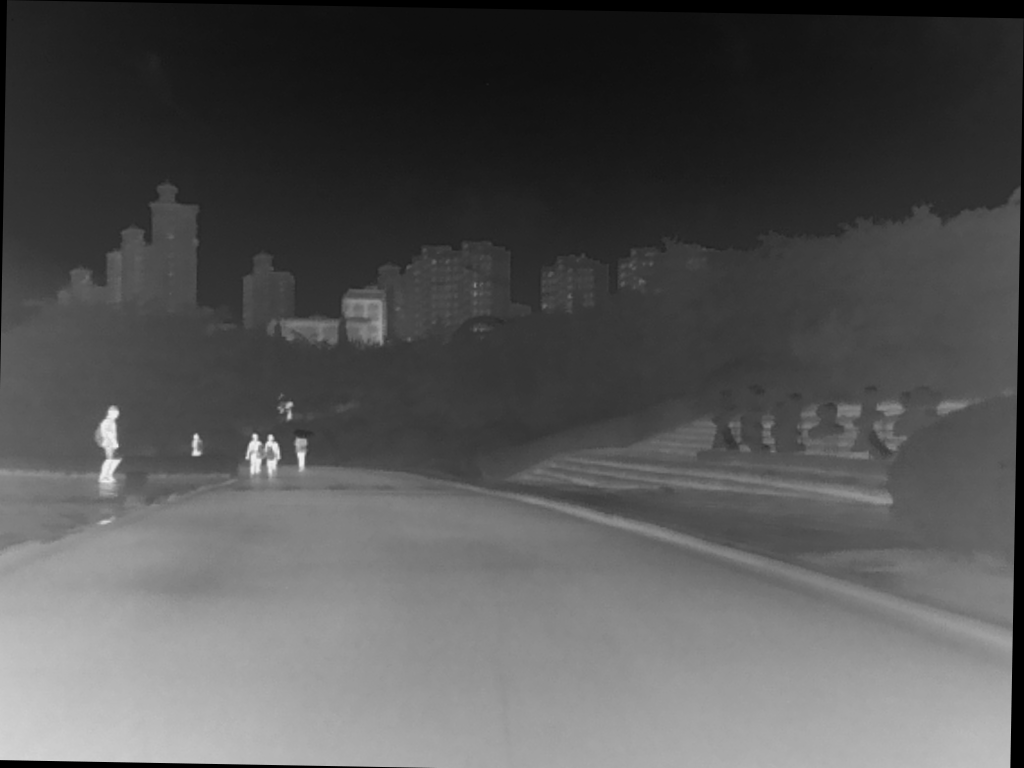} &
\includegraphics[width=.17\textwidth, height=2.0cm]{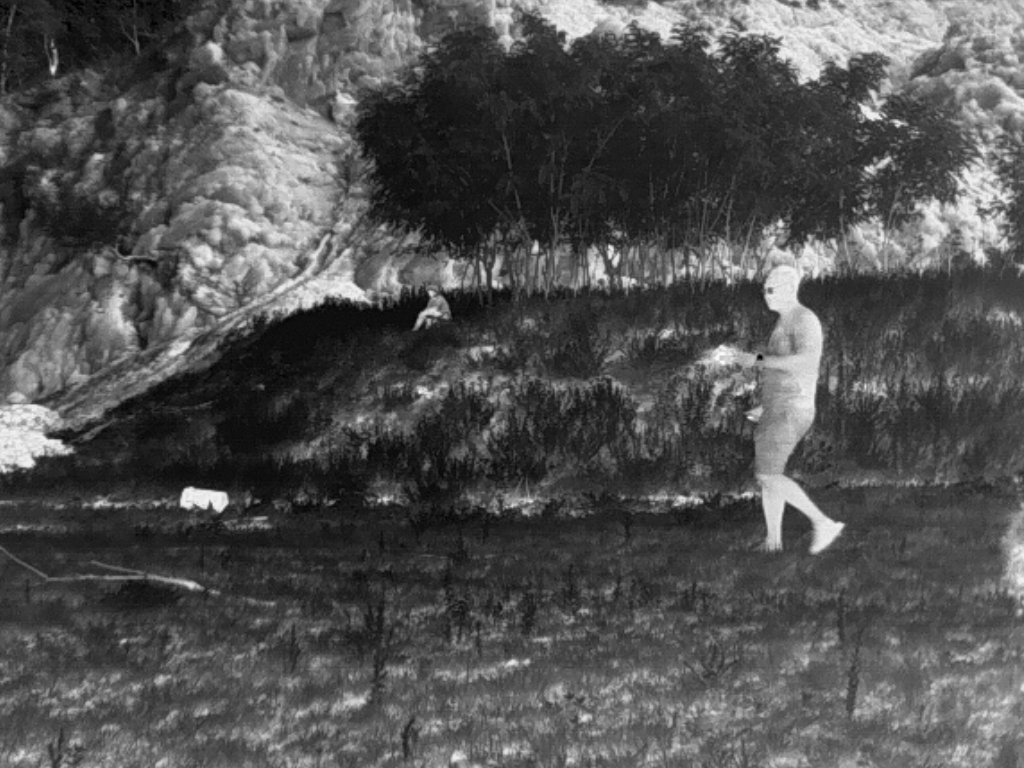} \\

\includegraphics[width=.17\textwidth, height=2.0cm]{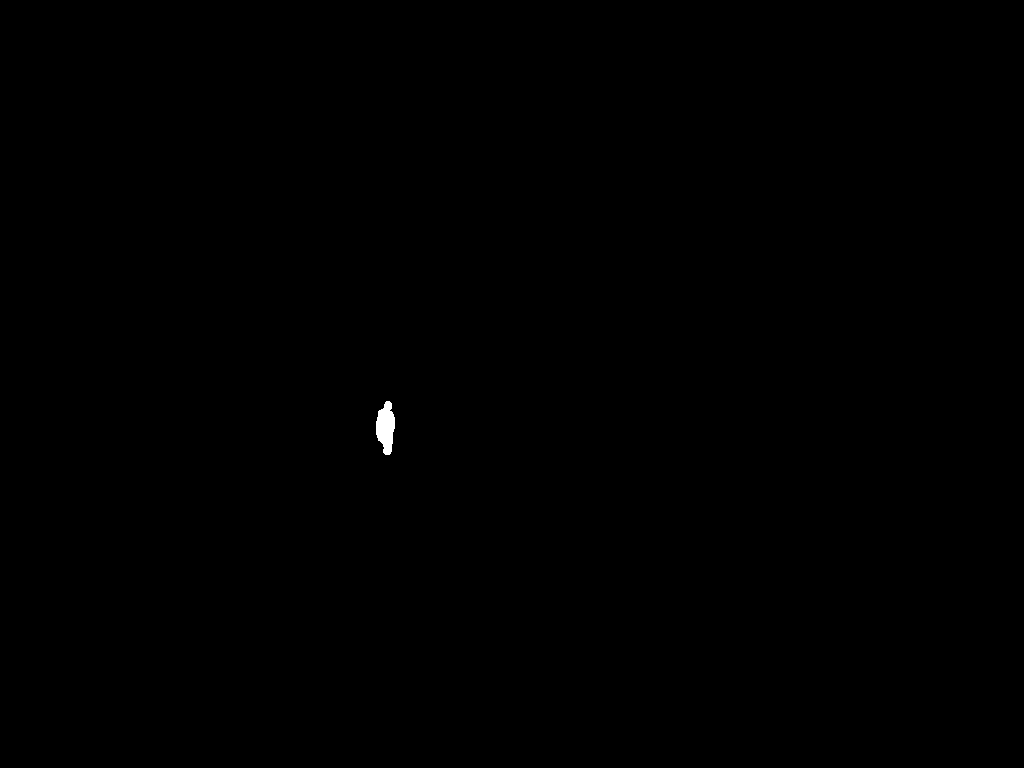} &
\includegraphics[width=.17\textwidth, height=2.0cm]{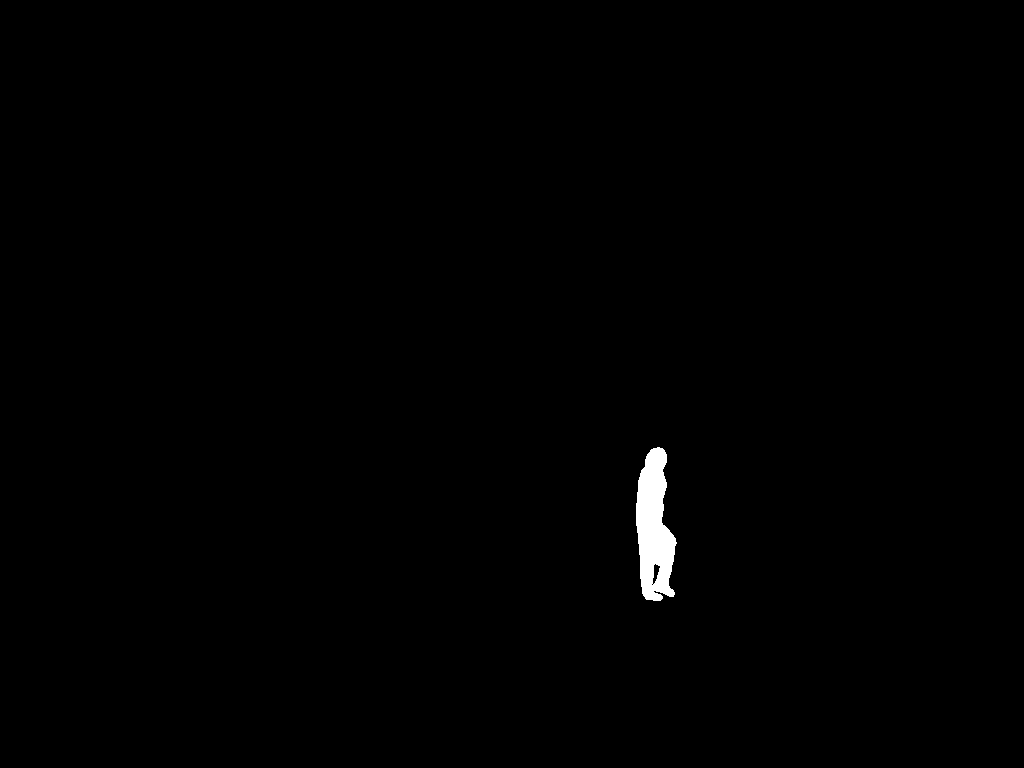} &
\includegraphics[width=.17\textwidth, height=2.0cm]{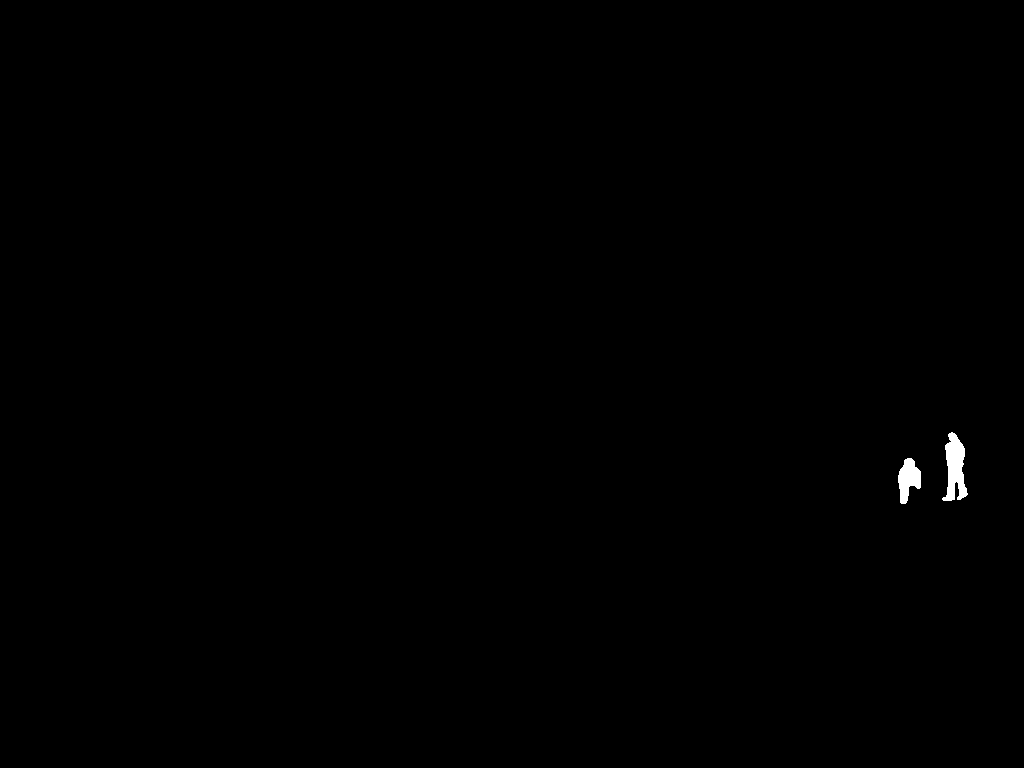} & 
\includegraphics[width=.17\textwidth, height=2.0cm]{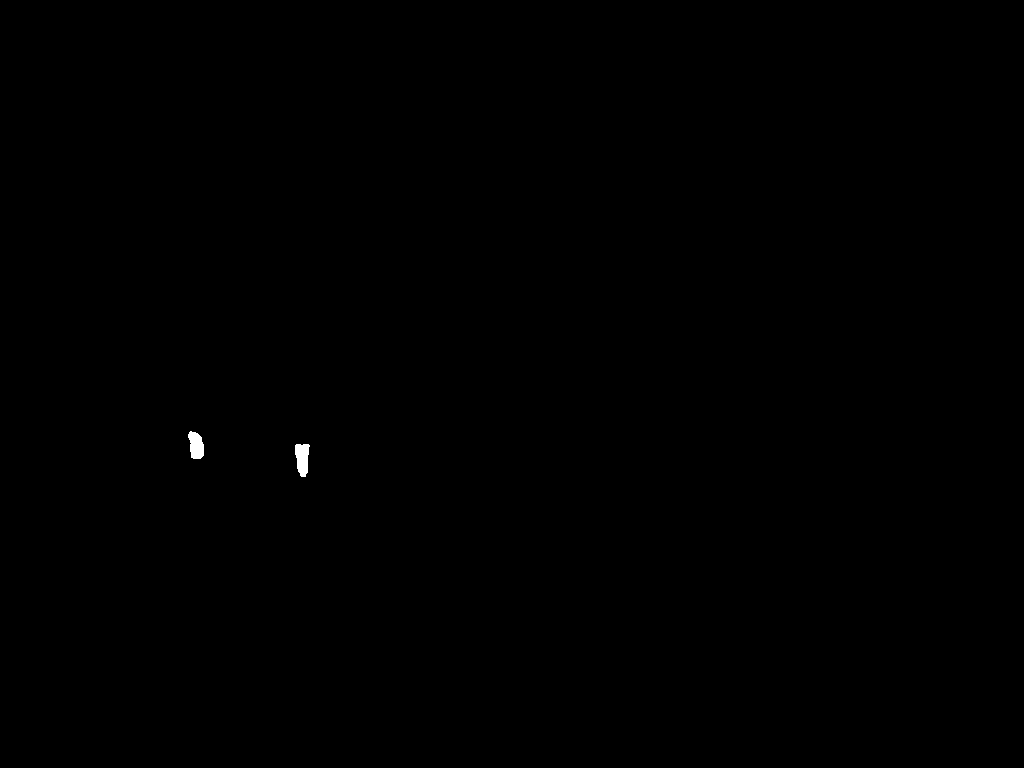} &
\includegraphics[width=.17\textwidth, height=2.0cm]{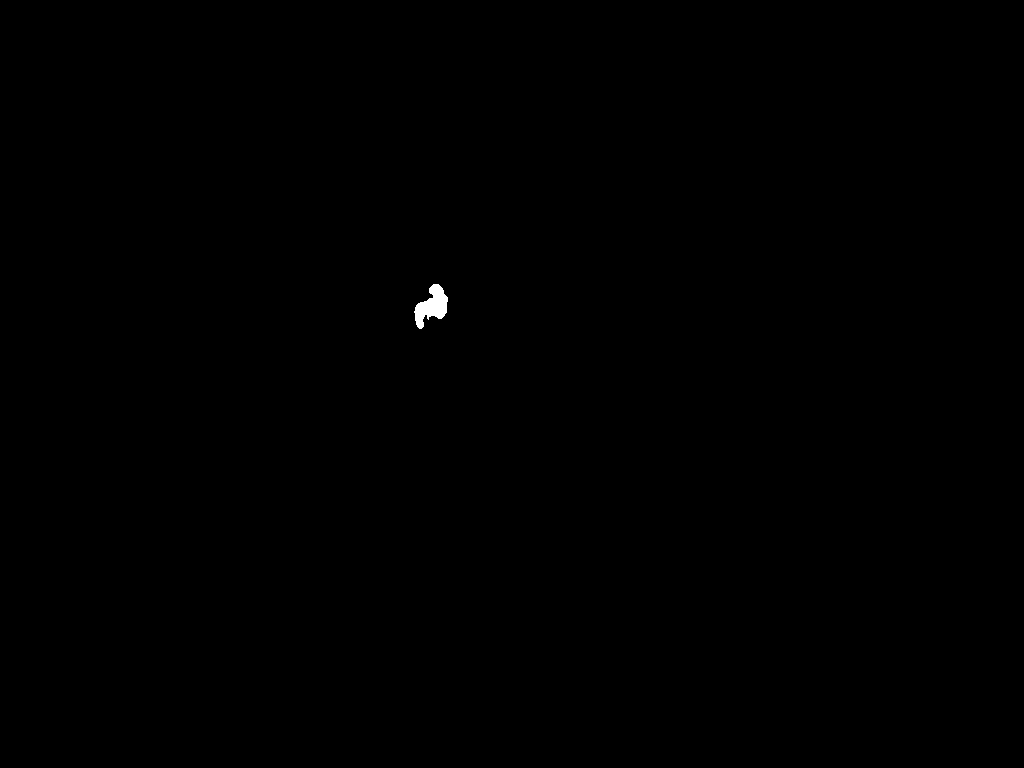} \\

\end{tabular}
}
\caption{Example images of the Camo-M3FD dataset. \textit{(1st row)} Visible (RGB) images. \textit{(2nd row)} Thermal images. \textit{(3rd row)} Segmentation mask images of camouflaged objects.}
\label{fig:dataset}
\end{figure*}

\section{Materials \& Methods}
\label{sec:prop}
This section delineates the systematic methodology employed in the curation and refinement of the Camo-M3FD dataset, a specialized benchmark derived from the M3FD dataset \cite{liu2022target} for camouflaged pedestrian detection. Figure~\ref{fig:dataset} shows example images of the final dataset obtained.

\begin{table*}
    \centering
    \begin{tabular}{cccccr}
         \toprule
         Dataset & Source & Year & Scope & Type of images & \# images \\
         \midrule
         Chameleon \cite{skurowski2018animal} & - & 2018 & Animal & RGB & 76 \\
         CAMO \cite{ltnghia-CVIU2019, Jinnan-IEEEAccess2021} & CVIU & 2019 & Animal \& others & RGB & 1,250 \\
         COD10K \cite{fan2020camouflaged, fan2021concealed} & CVPR & 2020 & Animal \& others & RGB & 10,000 \\
         NC4K \cite{yunqiu_cod21} & CVPR & 2021 & Animal \& others & RGB & 4,121 \\
         Camo-M3FD (Ours) & CVPR & 2026 & Pedestrian & RGB + Thermal & 614\\
         \bottomrule
    \end{tabular}
    \caption{COD datasets comparison.}
    \label{tab:dataset_comparison}
\end{table*}

\subsection{Data Collection}
The foundation of this work is the M3FD dataset \cite{liu2022target}, a state-of-the-art multispectral benchmark widely utilized for object detection tasks involving registered thermal and visible-spectrum imagery. While M3FD covers a broad range of urban scenarios (i.e., road, campus, street, harsh weather, disguise, haze, forest, and others) and different types of objects, the primary objective of Camo-M3FD is to address the specific challenge of camouflaged pedestrian detection, a critical yet underserved task in autonomous surveillance and security applications. To construct this subset, an exhaustive manual audit of the M3FD repository\footnote{https://github.com/dlut-dimt/TarDAL} is performed, isolating only those frames containing pedestrian instances. This focused selection ensures that the resulting dataset provides a rigorous testbed for detecting human targets that exhibit high degrees of visual blending with their environmental surroundings.

\subsection{Camouflage Quantification}
Following the selection of pedestrian-centric imagery and because M3FD only contains bounding boxes, high-precision mask annotations are conducted using the CVAT (Computer Vision Annotation Tool)\footnote{https://www.cvat.ai/}. After making annotations on all the images, a rigorous filtering step is implemented to quantitatively define the "camouflage level" of each instance, ensuring the dataset's integrity as a specialized benchmark.

The quantification of camouflage levels is grounded in the tripartite metric framework proposed by Lamdouar et al. \cite{lamdouar2023making}, which evaluates the visual relationship between a target and its immediate surroundings through color, texture, and structural coherence. The first component, Color Receptive Similarity ($S^{Q}_{rf}$), assesses the spectral alignment by comparing color distributions and intensities between the foreground and background. The second, Texture/Boundary Similarity ($S^{Q}_{b}$), measures the continuity of spatial patterns and the absence of disruptive edge gradients, determining how well the object's surface patterns blend into the environmental context. Finally, the Combined Camouflage Score ($S^{Q}_{\alpha}$) acts as a holistic descriptor by performing a weighted fusion of $S_{rf}$ and $S_{b}$, providing a single robust value that indicates total cryptic efficacy. 

The $S_{\alpha}$ score is calculated for every annotated instance in the initial pool. To establish a robust threshold that accounts for the inherent distribution of the data, the median and standard deviation of the population are utilized. An image is classified as "valid camouflaged data" only if the target meets the following statistical criterion:

\begin{equation}
S^{Q}_{\alpha} \geq \text{Median}(S^{Q}_{\alpha}) - \sigma(S^{Q}_{\alpha}),
\end{equation}

\noindent where $\sigma$ denotes the standard deviation. This approach allows for the filtration of outliers—specifically, highly conspicuous targets—while maintaining a diverse range of challenging, naturally camouflaged scenarios.

\subsection{Data Statistics and Selection}
\begin{figure}
    \centering
    \includegraphics[width=1\linewidth]{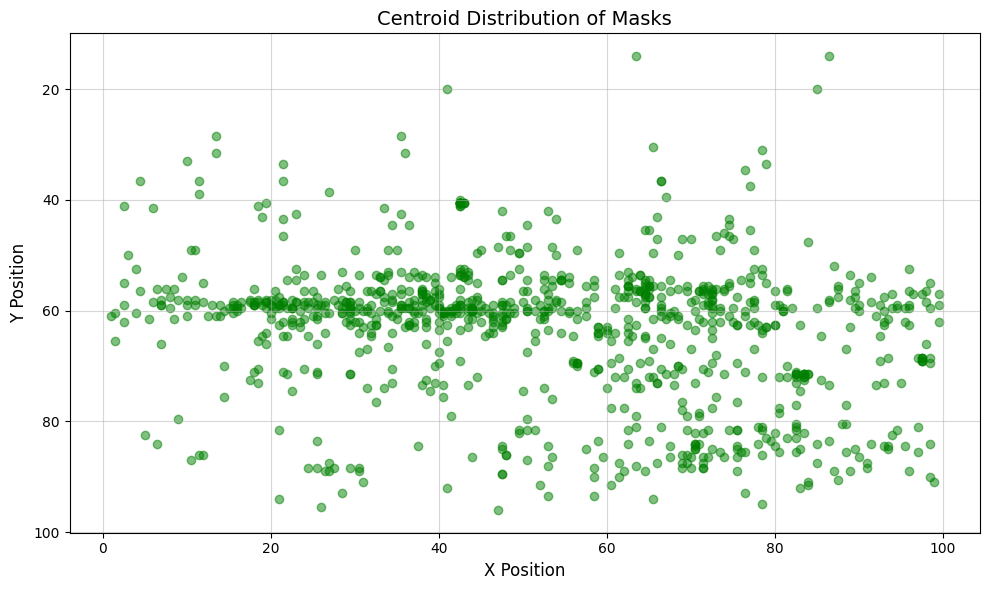}
    \caption{Spatial distribution of the centroids of the annotated GT masks.}
    \label{fig:mask_centroid}
\end{figure}

\begin{figure}
    \centering
    \includegraphics[width=1\linewidth]{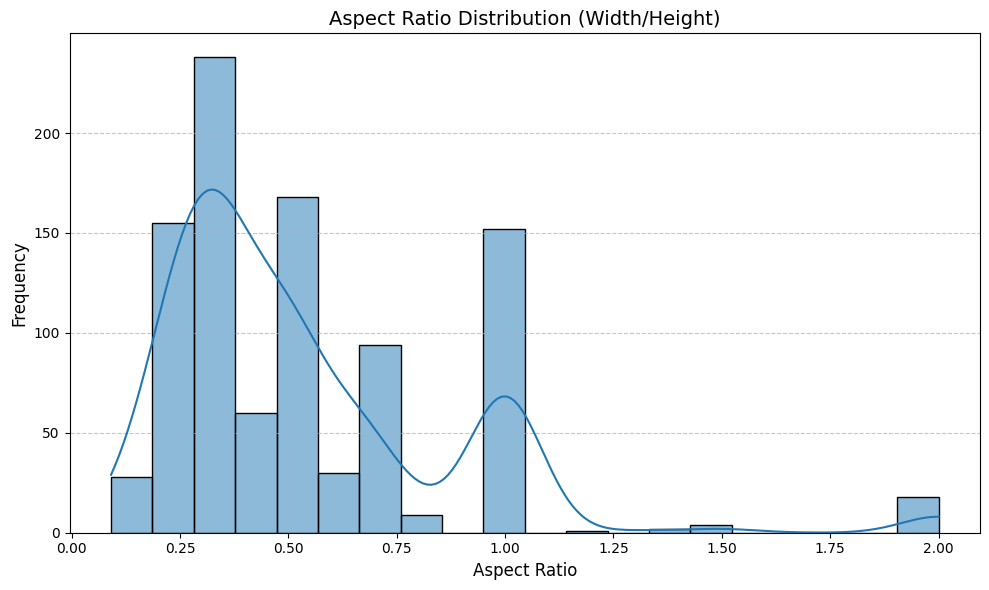}
    \caption{Aspect-ratio distribution of the GT masks.}
    \label{fig:aspect_ratio_mask}
\end{figure}

\begin{figure*}[!h]
\setlength\tabcolsep{0.75pt}
\centering
\scalebox{1.0}{
\begin{tabular}{cccccccccccccc}

\rotatebox{90}{RGB} & 
\includegraphics[width=.08\textwidth,height=2.15cm]{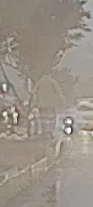} & 
\includegraphics[width=.08\textwidth,height=2.15cm]{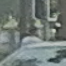} & 
\includegraphics[width=.08\textwidth,height=2.15cm]{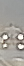} &
\includegraphics[width=.08\textwidth,height=2.15cm]{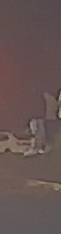} &
\includegraphics[width=.08\textwidth,height=2.15cm]{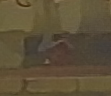} &
\includegraphics[width=.08\textwidth,height=2.15cm]{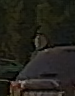} & 
\includegraphics[width=.08\textwidth,height=2.15cm]{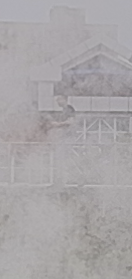} &
\includegraphics[width=.08\textwidth,height=2.15cm]{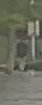} &
\includegraphics[width=.08\textwidth,height=2.15cm]{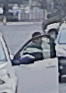} &
\includegraphics[width=.08\textwidth,height=2.15cm]{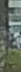} &
\includegraphics[width=.08\textwidth,height=2.15cm]{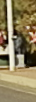} \\

\rotatebox{90}{RGB Sobel} & 
\includegraphics[width=.08\textwidth,height=2.15cm]{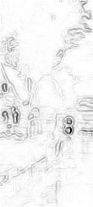} & 
\includegraphics[width=.08\textwidth,height=2.15cm]{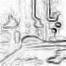} & 
\includegraphics[width=.08\textwidth,height=2.15cm]{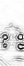} &
\includegraphics[width=.08\textwidth,height=2.15cm]{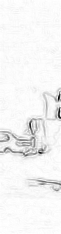} &
\includegraphics[width=.08\textwidth,height=2.15cm]{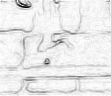} &
\includegraphics[width=.08\textwidth,height=2.15cm]{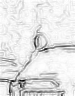} & 
\includegraphics[width=.08\textwidth,height=2.15cm]{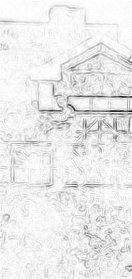} &
\includegraphics[width=.08\textwidth,height=2.15cm]{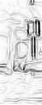} &
\includegraphics[width=.08\textwidth,height=2.15cm]{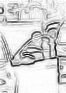} &
\includegraphics[width=.08\textwidth,height=2.15cm]{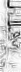} &
\includegraphics[width=.08\textwidth,height=2.15cm]{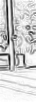} \\

\rotatebox{90}{GT Edges} & 
\includegraphics[width=.08\textwidth,height=2.15cm]{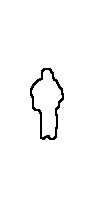} & 
\includegraphics[width=.08\textwidth,height=2.15cm]{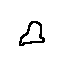} & 
\includegraphics[width=.08\textwidth,height=2.15cm]{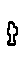} &
\includegraphics[width=.08\textwidth,height=2.15cm]{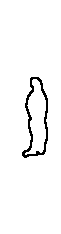} &
\includegraphics[width=.08\textwidth,height=2.15cm]{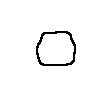} &
\includegraphics[width=.08\textwidth,height=2.15cm]{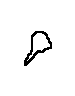} &
\includegraphics[width=.08\textwidth,height=2.15cm]{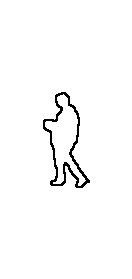} &
\includegraphics[width=.08\textwidth,height=2.15cm]{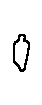} &
\includegraphics[width=.08\textwidth,height=2.15cm]{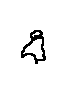} &
\includegraphics[width=.08\textwidth,height=2.15cm]{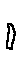} &
\includegraphics[width=.08\textwidth,height=2.15cm]{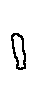} \\


$S_{\alpha}$ & 0.8059 & 0.6292 & 0.3946 & 0.7715 & 0.7711 & 0.4472 & 0.7873 & 0.6052 & \cred{0.3358} & \cred{0.3573} & \cred{0.3600} \\

\end{tabular}
}
\caption{Examples of accepted and rejected (marked in red) images alongside their respective edges extracted by RGB using Sobel, edges of the GT mask, and camouflage scores ($S_{\alpha}$).}
\label{fig:cameval}
\end{figure*}

To provide a comprehensive overview of the dataset's characteristics, the spatial and geometric properties of the annotations are analyzed. Figure~\ref{fig:mask_centroid} illustrates the spatial distribution of the centroids of the annotated GT masks, demonstrating a varied coverage across the image plane. Figure~\ref{fig:aspect_ratio_mask} presents the aspect ratio distribution of the GT masks, highlighting the diversity in pedestrian poses and scales captured. Finally, Figure~\ref{fig:cameval} provides a qualitative comparison of the selection process, showcasing examples of accepted and rejected images alongside their RGB-Sobel edges, the GT mask edges, and camouflage scores to validate the efficacy of the filtering threshold. 

The statistical analysis of the initial dataset yielded a median $S^{Q}_{rf}$ of $0.4636$ with a standard deviation ($\sigma$) of $0.2881$, while the $S_{b}$ exhibited a median of $0.6130$ and a lower dispersion of $\sigma = 0.0818$. The primary filtering metric, the Combined Camouflage Score ($S_{\alpha}$), yielded a median of $0.5292$ and a standard deviation of $0.1669$. Based on these population statistics, the acceptance threshold is established as $\text{Median}(S^{Q}_{\alpha}) - \sigma(S^{Q}_{\alpha})$, defining a rigorous accepted range of $[0.3623, 1.0000]$ for the final dataset inclusion.

The rigorous filtering process resulted in a final curated set of 614 valid RGB-T image pairs of camouflaged pedestrians. This refined dataset is partitioned into training, validation, and testing sets using an 80/10/10 ratio. Specifically, 492 image pairs are used for training, while 61 and 61 pairs are reserved for validation and testing, respectively, ensuring a balanced distribution for model development and unbiased performance evaluation.

\begin{table*}[!h]
    \centering
    \caption{Distinctive characteristics of the evaluated SoTA COD techniques.}
    \begin{tabular}{l|cccclr}
        \hline
        \textbf{Technique} & \textbf{Source} & \textbf{Source} & \textbf{Year} & \textbf{Image Size} & \textbf{Backbone} & \textbf{\#Param.} \\
        & & \textbf{Type} & & \textbf{(px)} & & \textbf{(M)} \\
        \hline
        BASNet \cite{qin2019basnet} & CVPR & Conference & 2019 & $256 \times 256$ & ResNet-34 \cite{he2016deep} & 87.06 \\
        SINet-v2 \cite{fan2021concealed} & TPAMI & Journal & 2021 & $352 \times 352$ & Res2Net-50 \cite{gao2019res2net} & 24.93 \\       
        BGNet \cite{chen2022boundary} & IJCAI & Conference & 2022 & $416 \times 416$ & Res2Net-50 \cite{gao2019res2net} & 77.80 \\
        C$^{2}$F-Net \cite{chen2022camouflaged} & TCSVT & Conference & 2022 & $352 \times 352$ & Res2Net-50 \cite{gao2019res2net} & 26.36 \\
        OCENet \cite{liu2022modeling} & WACV & Conference & 2022 & $352 \times 352$ & ResNet-50 \cite{he2016deep} & 58.17 \\
        EAMNet \cite{sun2023edge} & ICME & Conference & 2023 & $384 \times 384$ & Res2Net-50 \cite{gao2019res2net} & 30.51 \\
        DGNet \cite{ji2023deep} & MIR & Journal & 2023 & $352 \times 352$ & EfficientNet \cite{tan2019efficientnet} & 8.30 \\
        HitNet \cite{hu2023high} & AAAI & Conference & 2023 & $352 \times 352$ & PVTv2 \cite{wang2022pvt} & 25.73 \\
        PCNet \cite{yang2024plantcamo} & arXiv & - & 2024 & $352 \times 352$ & PVTv2 \cite{wang2022pvt} & 27.66 \\
        CTF-Net \cite{zhang2025effective} & CVIU & Journal & 2025 & $384 \times 384$ & PVTv2 \cite{wang2022pvt} & 64.48 \\ 
        AVNet \cite{velesaca2026iguana} & VISAPP & Conference & 2026 & $416 \times 416$ & PVTv2 \cite{wang2022pvt} & 48.04 \\
        
        \hline
    \end{tabular}
    \label{tab:networks_eval}
\end{table*}

\subsection{Evaluated Architectures}
To benchmark the Camo-M3FD dataset, a representative selection of state-of-the-art (SoTA) models was evaluated, encompassing diverse architectural strategies for Camouflaged Object Detection (COD). These models are categorized based on their approach to boundary preservation and feature integration. Explicit edge-modeling networks, such as SINet \cite{fan2021concealed}, EAMNet \cite{sun2023edge}, BGNet \cite{chen2022boundary}, and DGNet \cite{ji2023deep}, are utilized to assess the efficacy of dedicated modules in capturing subtle contrast variations and object contours.

In parallel, implicit refinement frameworks are evaluated, including those focusing on hybrid losses (BASNet \cite{qin2019basnet}), iterative feedback (HitNet \cite{hu2023high}), and uncertainty modeling (OCENet \cite{liu2022modeling}). Global-local context integration is assessed through Transformer-based and fusion-centric models like CTF-Net \cite{zhang2025effective}, C$^{2}$F-Net \cite{chen2022camouflaged}, and CHNet \cite{wang2025efficient}. Furthermore, specialized channel-interaction networks, notably ARNet \cite{wang2025assisted} and ARNet-V2 \cite{wang2025assistedv2}, are included to measure the impact of sophisticated feature refinement. Finally, the benchmark incorporates domain-specific and multimodal architectures, such as the plant-targeted PCNet \cite{yang2024plantcamo} and the cross-spectral AVNet \cite{velesaca2026iguana}, to evaluate performance across diverse ecological and multimodal scenarios. A comprehensive summary of these architectural characteristics is provided in Table~\ref{tab:networks_eval}.

\subsection{Metrics} 
\label{subsec:metrics} 
The performance of the SoTA COD approaches mentioned in the previous section is rigorously evaluated using five standard quantitative metrics, ensuring a holistic assessment of model accuracy and robustness. These metrics include: Structure-measure ($S_\alpha$) \cite{fan2017structure}, weighted F-measure ($F^w_\beta$) \cite{margolin2014evaluate}, Mean Absolute Error ($M$) \cite{perazzi2012saliency}, Enhanced-alignment measure ($E_\phi$) \cite{fan2018enhanced}, and the traditional F-measure ($F_\beta$) \cite{achanta2009frequency}. The $S_\alpha$ metric is utilized to quantify structural similarity by evaluating both region-aware and object-aware correlations between the prediction and ground truth, thereby measuring the preservation of global structural integrity. To address the limitations of pixel-wise comparisons, the $F^w_\beta$ incorporates spatial weights to provide an improved assessment of segmentation quality, emphasizing boundary precision and the spatial distribution of errors. For pixel-level accuracy, the $M$ metric calculates the average absolute difference between the normalized saliency maps and the binary ground truth. Furthermore, the $E_\phi$ metric leverages human visual perception mechanisms to simultaneously evaluate local pixel matching and global image statistics. Lastly, $F_\beta$ offers a harmonic mean of precision and recall, serving as a fundamental measure of overall detection efficacy. To capture the performance across varying confidence levels, multiple variants of the F-measure and E-measure are computed. This includes the adaptive threshold version ($F^{adp}_\beta$, $E^{adp}_\phi$), as well as the mean ($F^{mean}_\beta$, $E^{mean}_\phi$) and maximum ($F^{max}_\beta$, $E^{max}_\phi$) values derived from the precision-recall curves. This multi-faceted evaluation strategy ensures that the proposed framework is benchmarked against both deterministic and threshold-agnostic performance criteria.

\begin{table*}[!h]
    \centering
    \caption{Metric evaluation results for each COD technique on the Camo-M3FD dataset, reported for the RGB and Thermal baseline. Results are presented using the metric notation defined in Sec.~\ref{subsec:metrics}, ``$\uparrow/\downarrow$'' indicates that larger or smaller is better. The best three performing results are highlighted using color: \First{First}, \Second{Second}, and \Third{Third} respectively.}
    \resizebox{2\columnwidth}{!}{
    \begin{tabular}{l|c|rrrrrrrrr}
        \toprule
        Technique & Input & $S_\alpha \uparrow$ & $F^{w}_\beta \uparrow$ & $M \downarrow$ & $E^{adp}_\phi \uparrow$ & $E^{mean}_\phi \uparrow$ & $E^{max}_\phi \uparrow$ & $F^{adp}_\beta \uparrow$ & $F^{mean}_\beta \uparrow$ & $F^{max}_\beta \uparrow$ \\
        
        \midrule

        \multirow{2}{*}{BASNet \cite{qin2019basnet}} & Vis & 0.6239 & 0.2902 & 0.0032 & 0.6972 & 0.7183 & 0.8042 & 0.2879 & 0.3057 & 0.3137 \\
        & Th & 0.7051 & 0.4161 & \second{0.0028} & 0.7293 & 0.7822 & 0.8078 & 0.3762 & 0.4358 & 0.4571 \\ 

        \midrule
        
        \multirow{2}{*}{SINet-v2 \cite{fan2021concealed}} & Vis & 0.6275 & 0.2693 & 0.0037 & 0.6039 & 0.7080 & 0.7227 & 0.2244 & 0.2872 & 0.3033 \\
        & Th & 0.6927 & 0.4072 & 0.0034 & 0.6450 & 0.7593 & 0.7949 & 0.3428 & 0.4244 & 0.4424 \\ 
        
        \midrule

        \multirow{2}{*}{BGNet \cite{chen2022boundary}} & Vis & 0.6745 & 0.3922 & 0.0500 & 0.7594 & 0.7687 & 0.8142 & 0.3576 & 0.4124 & 0.4255 \\
        & Th & 0.7196 & 0.4699 & 0.0106 & 0.7664 & \first{0.8306} & \third{0.8539} & \third{0.4315} & 0.4865 & 0.4963 \\
                
        \midrule

        \multirow{2}{*}{C$^{2}$F-Net \cite{chen2022camouflaged}} & Vis & 0.5137 & 0.0432 & 0.0811 & 0.4804 & 0.6079 & 0.7333 & 0.1433 & 0.2155 & 0.2554 \\ 
        & Th & 0.5244 & 0.0522 & 0.0663 & 0.5122 & 0.6432 & 0.7656 & 0.2064 & 0.2882 & 0.3437 \\
        
        \midrule

        \multirow{2}{*}{OCENet \cite{liu2022modeling}} & Vis & 0.5994 & 0.2357 & 0.0037 & 0.6680 & 0.7975 & 0.8201 & 0.2240 & 0.2546 & 0.2632 \\
        & Th & \third{0.7277} & \third{0.4884} & 0.0037 & 0.7122 & 0.8152 & \first{0.8666} & 0.4253 & \second{0.4998} & \first{0.5403} \\ 
        
        \midrule 
        
        \multirow{2}{*}{EAMNet \cite{sun2023edge}} & Vis & 0.5227 & 0.0494 & 0.0160 & 0.4048 & 0.6141 & 0.8109 & 0.0998 & 0.1752 & 0.2352 \\
        & Th & 0.5047 & 0.0333 & 0.0506 & 0.4946 & 0.6458 & 0.8091 & 0.1836 & 0.2622 & 0.3799 \\ 
        
        \midrule
        
        \multirow{2}{*}{DGNet \cite{ji2023deep}} & Vis & 0.6438 & 0.3109 & 0.0039 & 0.6720 & 0.7598 & 0.7739 & 0.2759 & 0.3235 & 0.3377 \\
        & Th & 0.6898 & 0.4073 & 0.0052 & 0.6765 & 0.7928 & 0.8227 & 0.3586 & 0.4244 & 0.4403 \\
        
        \midrule
        
        \multirow{2}{*}{HitNet \cite{hu2023high}} & Vis & 0.5659 & 0.1593 & 0.0030 & 0.7333 & 0.5685 & 0.7353 & 0.1815 & 0.1721 & 0.1809 \\
        & Th & 0.6682 & 0.3622 & \third{0.0029} & 0.7694 & 0.7466 & 0.7778 & 0.3910 & 0.3800 & 0.3919 \\

        \midrule

        \multirow{2}{*}{PCNet \cite{yang2024plantcamo}} & Vis & 0.6512 & 0.3227 & 0.0034 & 0.5048 & 0.7639 & 0.8069 & 0.1688 & 0.3464 & 0.3552 \\
        & Th & 0.7034 & 0.4260 & 0.0030 & 0.6187 & \third{0.8280} & 0.8428 & 0.2674 & 0.4504 & 0.4572 \\ 
        
        \midrule


        \multirow{2}{*}{CTF-Net \cite{zhang2025effective}} & Vis & 0.5077 & 0.0525 & 0.0755 & 0.4201 & 0.5912 & 0.7296 & 0.1322 & 0.2449 & 0.3146 \\ 
        & Th & 0.6532 & 0.2955 & 0.0116 & 0.4178 & 0.7515 & 0.8073 & 0.1409 & 0.4310 & 0.4794 \\


        \midrule
        
        \multirow{3}{*}{AVNet \cite{velesaca2026iguana}} & Vis & 0.6669 & 0.4035 & \third{0.0029} & \first{0.8294} & 0.8164 & 0.8287 & 0.3923 & 0.3985 & 0.4068 \\ 
        & Th & \second{0.7289} & \second{0.5066} & \first{0.0026} & \third{0.7989} & 0.8075 & 0.8242 & \second{0.4831} & \third{0.4926} & \third{0.5113} \\ 
        & Vis+Th & \first{0.7318} & \first{0.5301} & 0.0030 & \second{0.8167} & \second{0.8287} & \second{0.8617} & \first{0.5051} & \first{0.5139} & \second{0.5362} \\ 
 
        \bottomrule

    \end{tabular}
    }
    \label{tab:results_cod_camo-m3fd}
\end{table*}

\begin{figure*}[!h]
\setlength\tabcolsep{0.75pt}
\centering
\scalebox{1.0}{
\begin{tabular}{ccccccccc}

\rotatebox{90}{\scriptsize{RGB}} & \includegraphics[width=.16\textwidth, height=2.1cm]{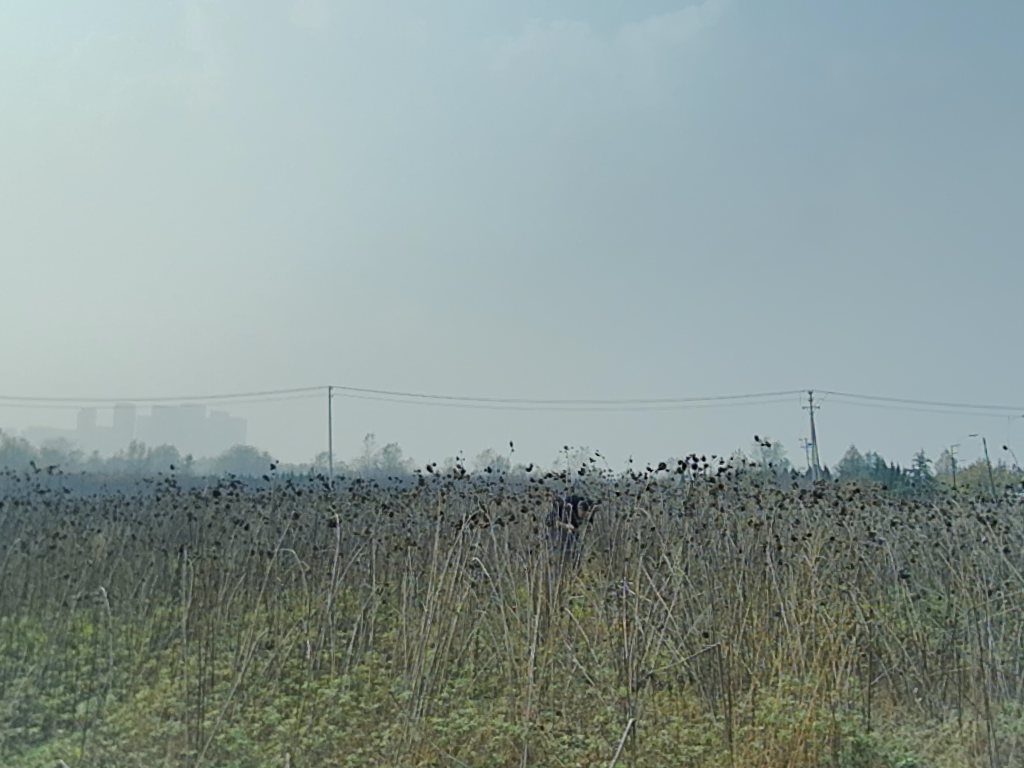} & 
\includegraphics[width=.16\textwidth, height=2.1cm]{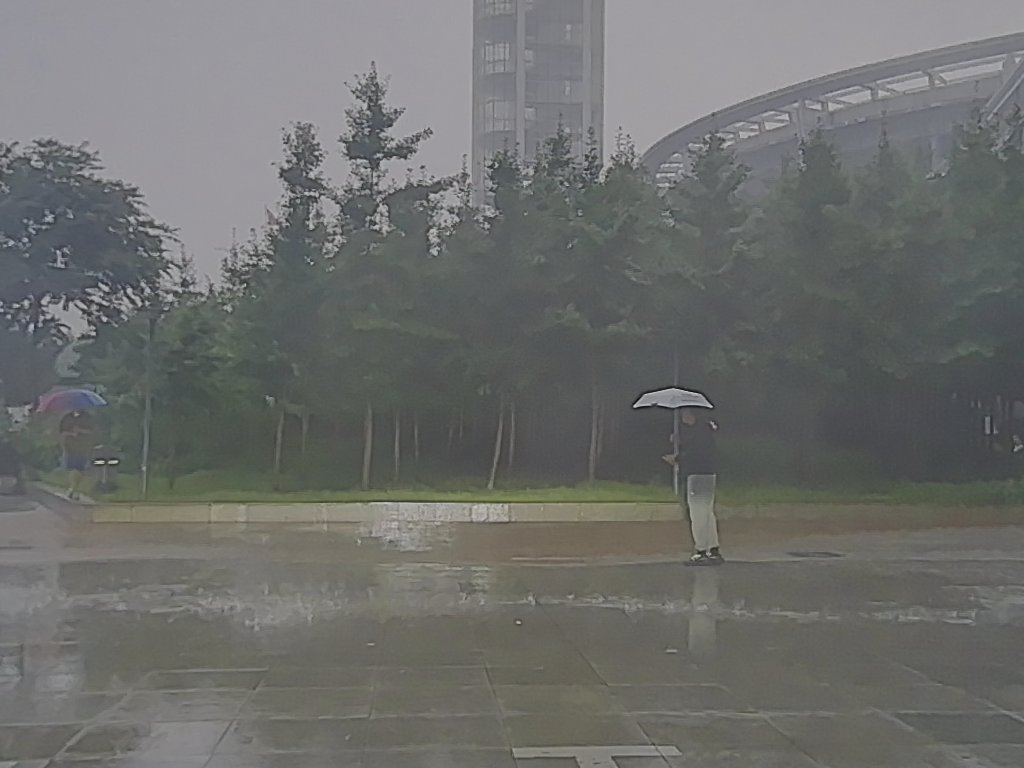} & 
\includegraphics[width=.16\textwidth, height=2.1cm]{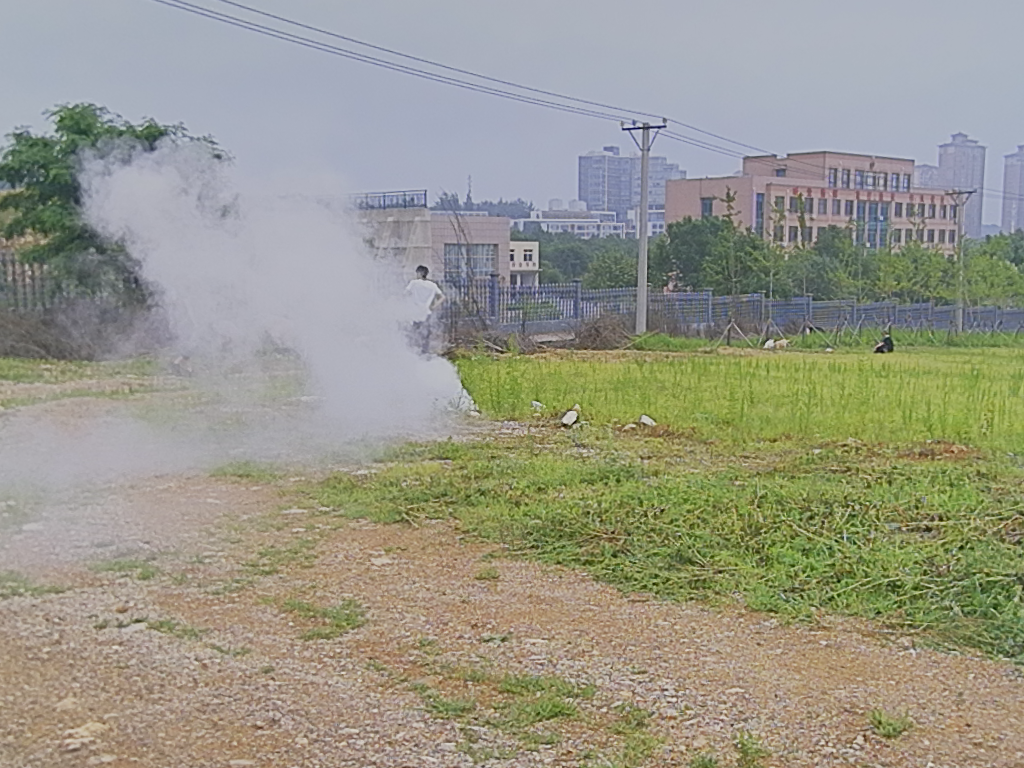} & 
\includegraphics[width=.16\textwidth, height=2.1cm]{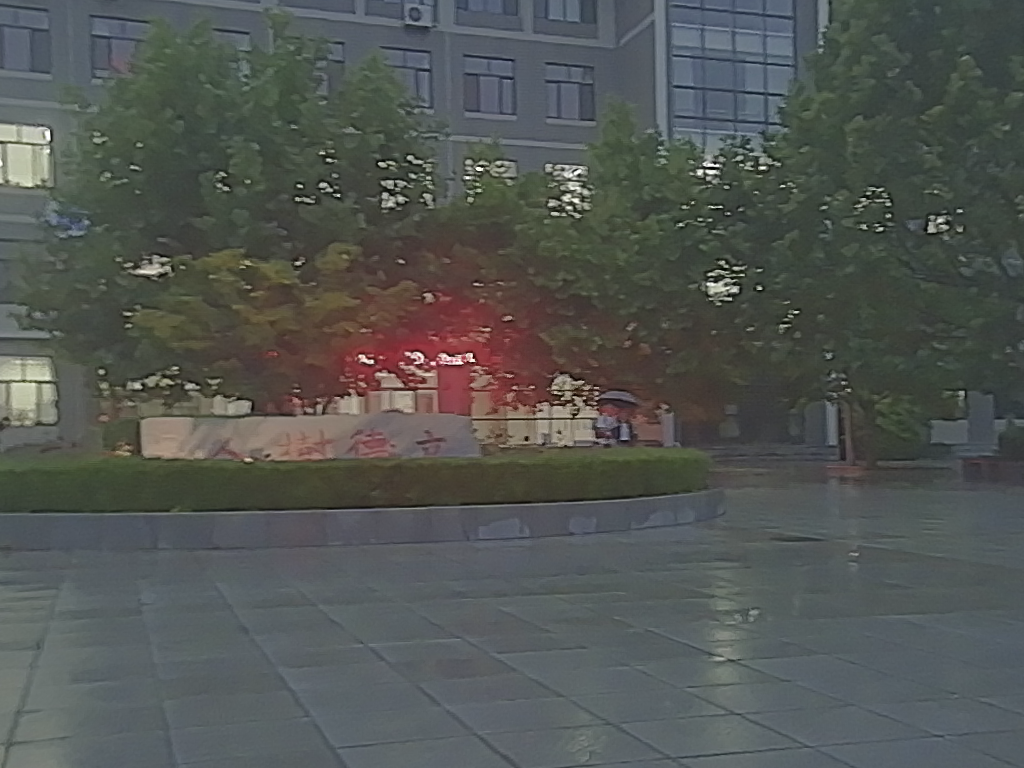} &  
\includegraphics[width=.16\textwidth, height=2.1cm]{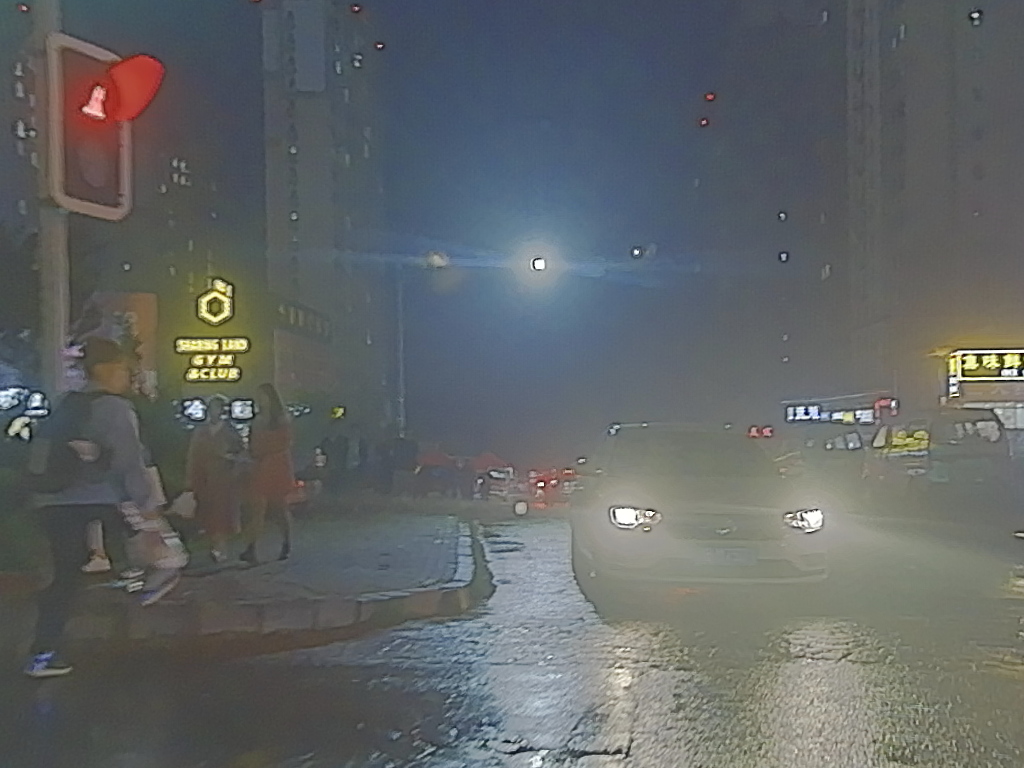} &  
\includegraphics[width=.16\textwidth, height=2.1cm]{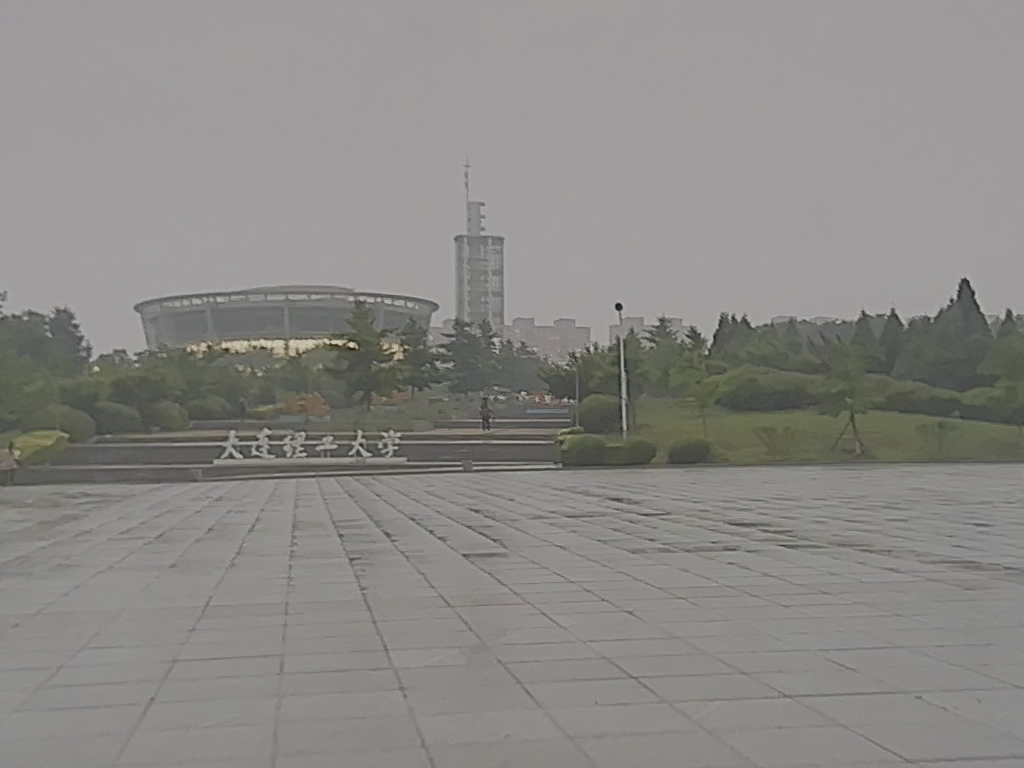} \\ 

\rotatebox{90}{\scriptsize{Thermal}} & \includegraphics[width=.16\textwidth, height=2.1cm]{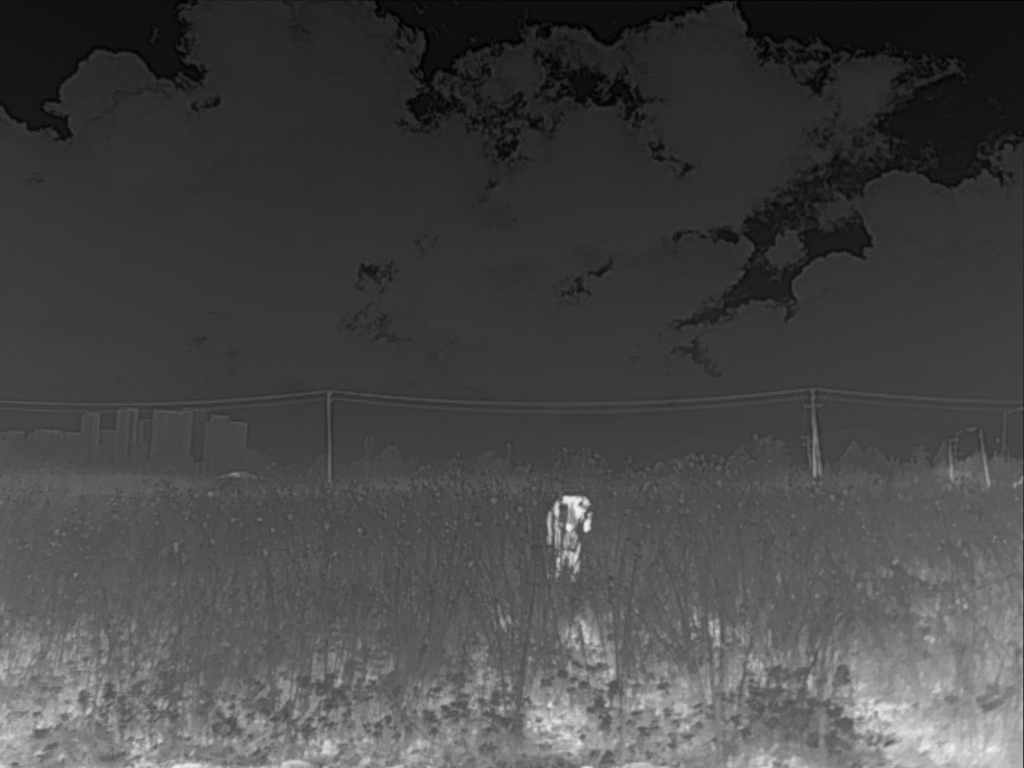} & 
\includegraphics[width=.16\textwidth, height=2.1cm]{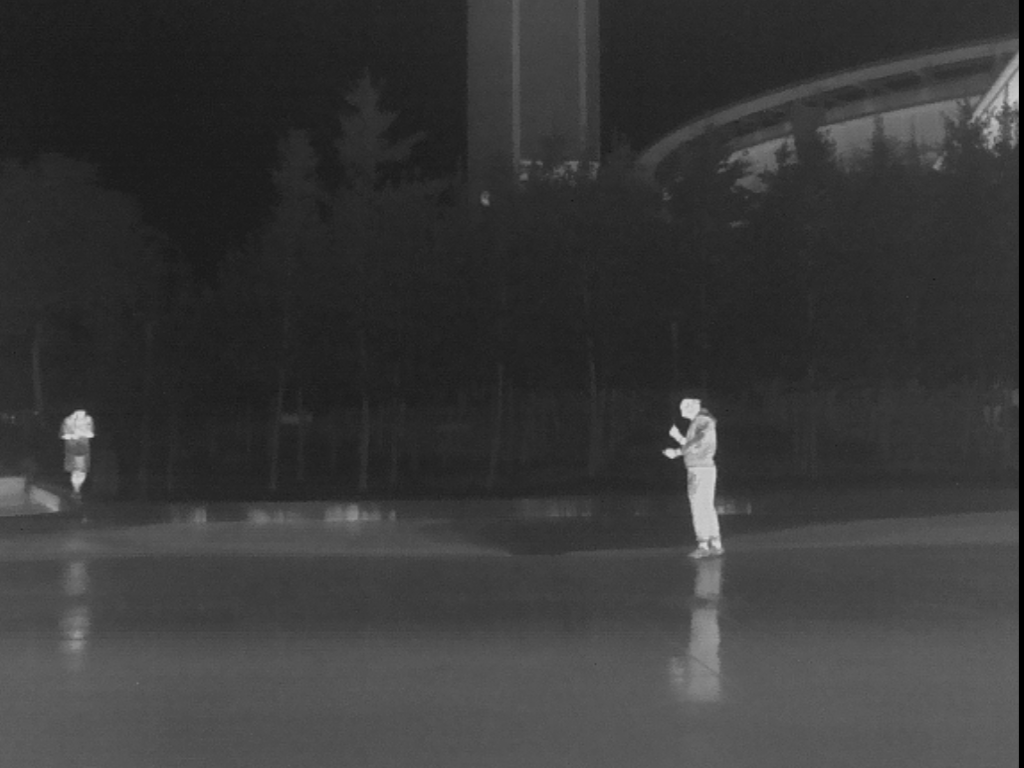} & 
\includegraphics[width=.16\textwidth, height=2.1cm]{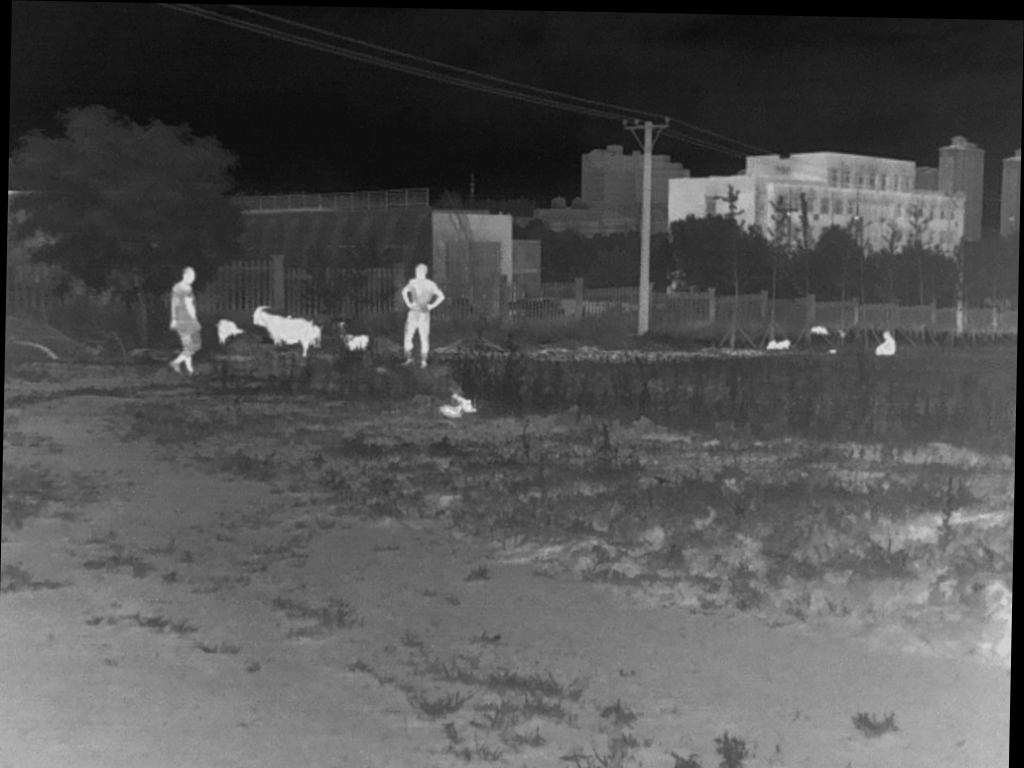} & 
\includegraphics[width=.16\textwidth, height=2.1cm]{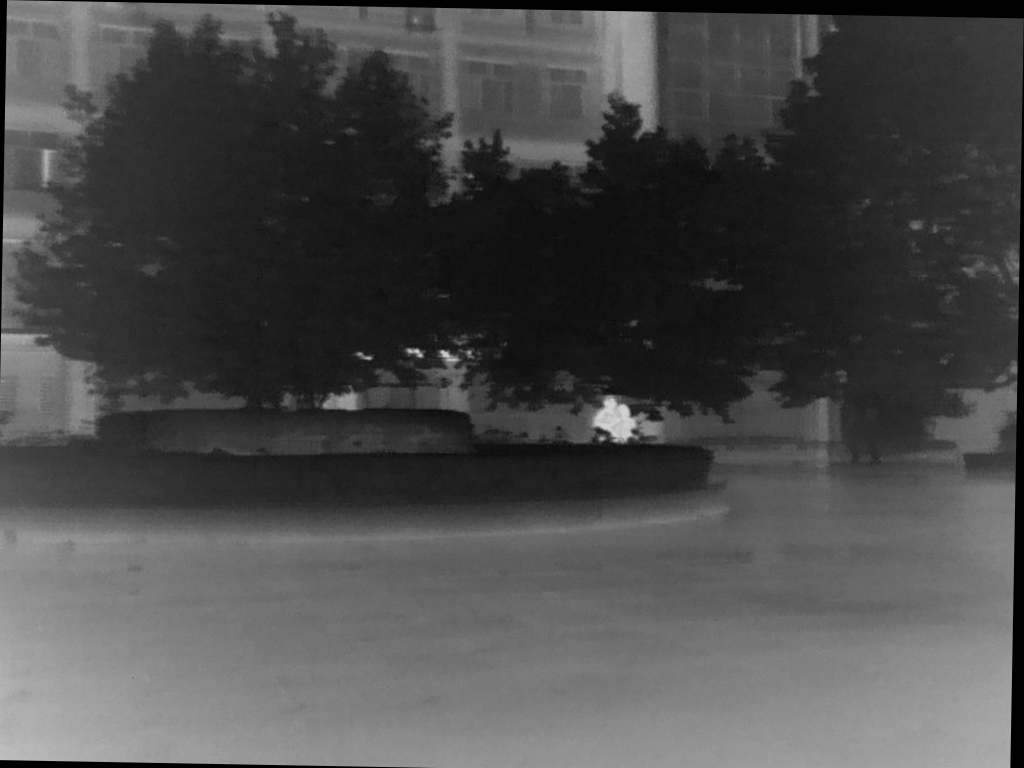} &  
\includegraphics[width=.16\textwidth, height=2.1cm]{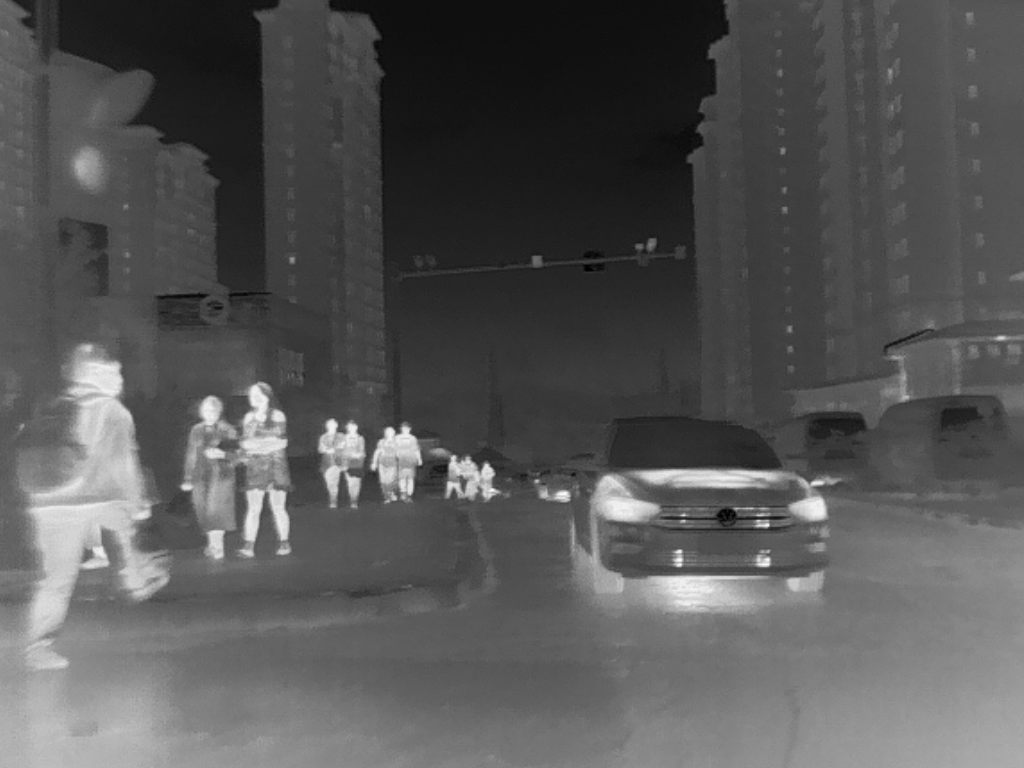} &  
\includegraphics[width=.16\textwidth, height=2.1cm]{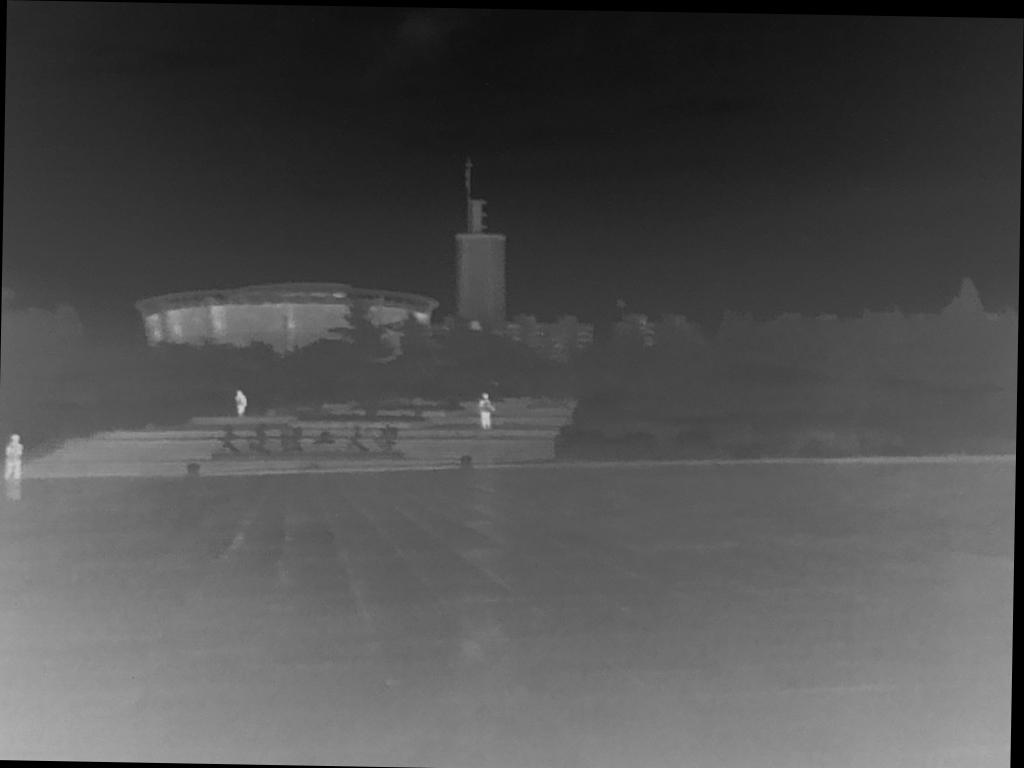} \\  

\rotatebox{90}{\scriptsize{GT}} & \includegraphics[width=.16\textwidth, height=2.1cm]{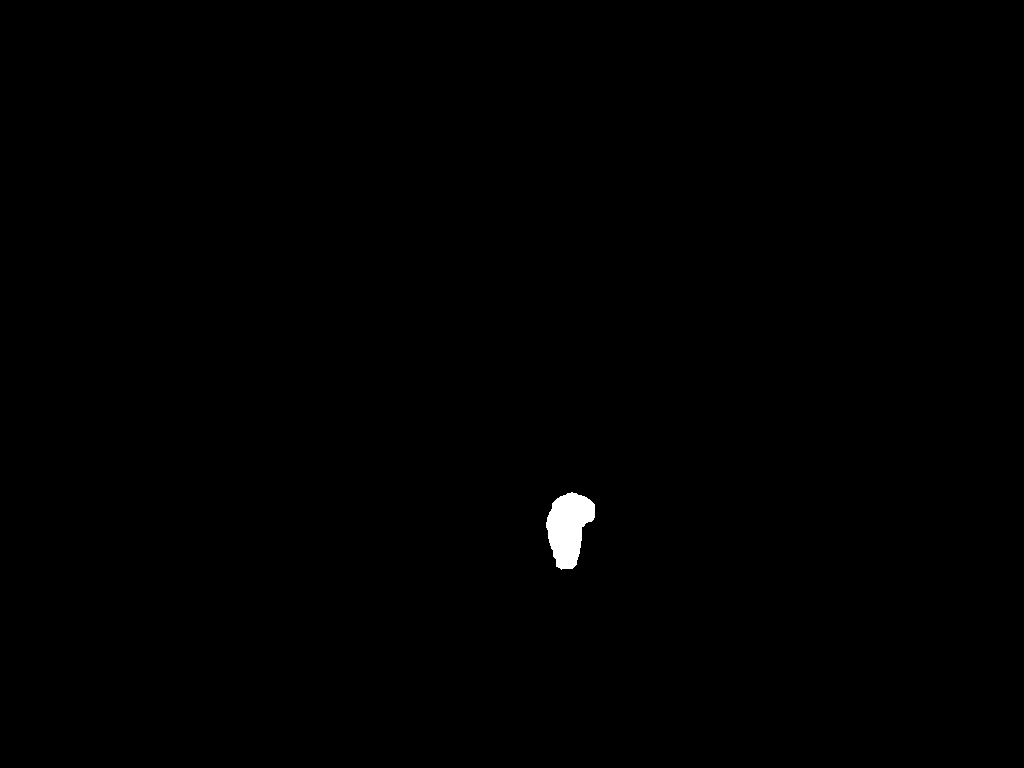} & 
\includegraphics[width=.16\textwidth, height=2.1cm]{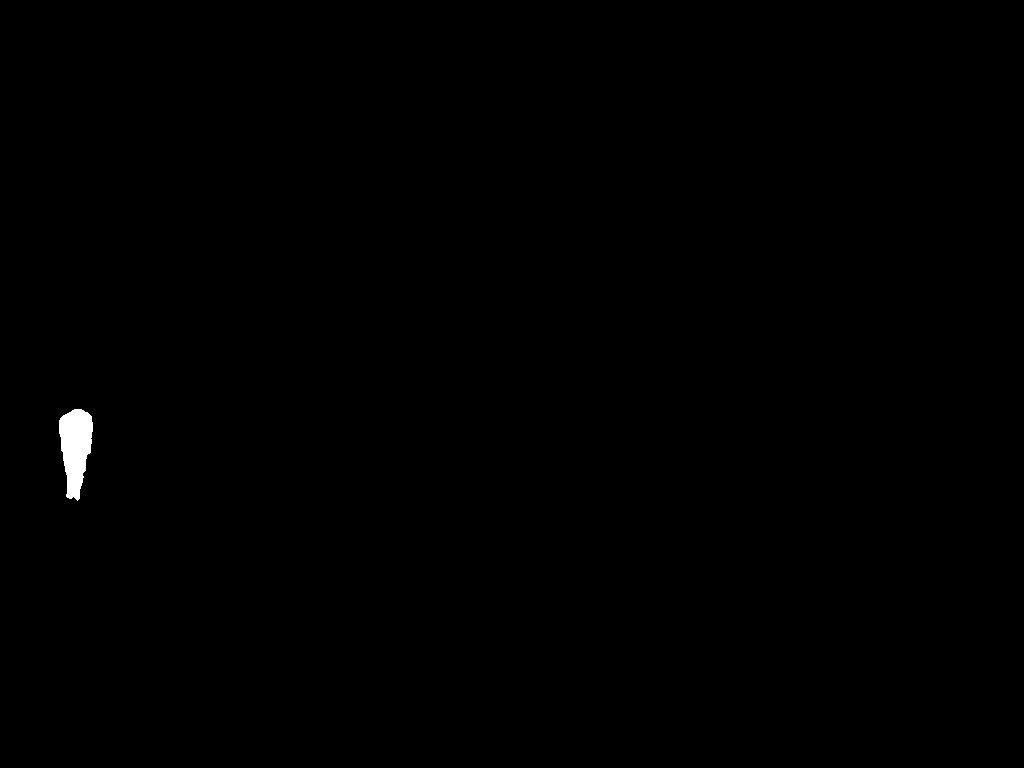} & 
\includegraphics[width=.16\textwidth, height=2.1cm]{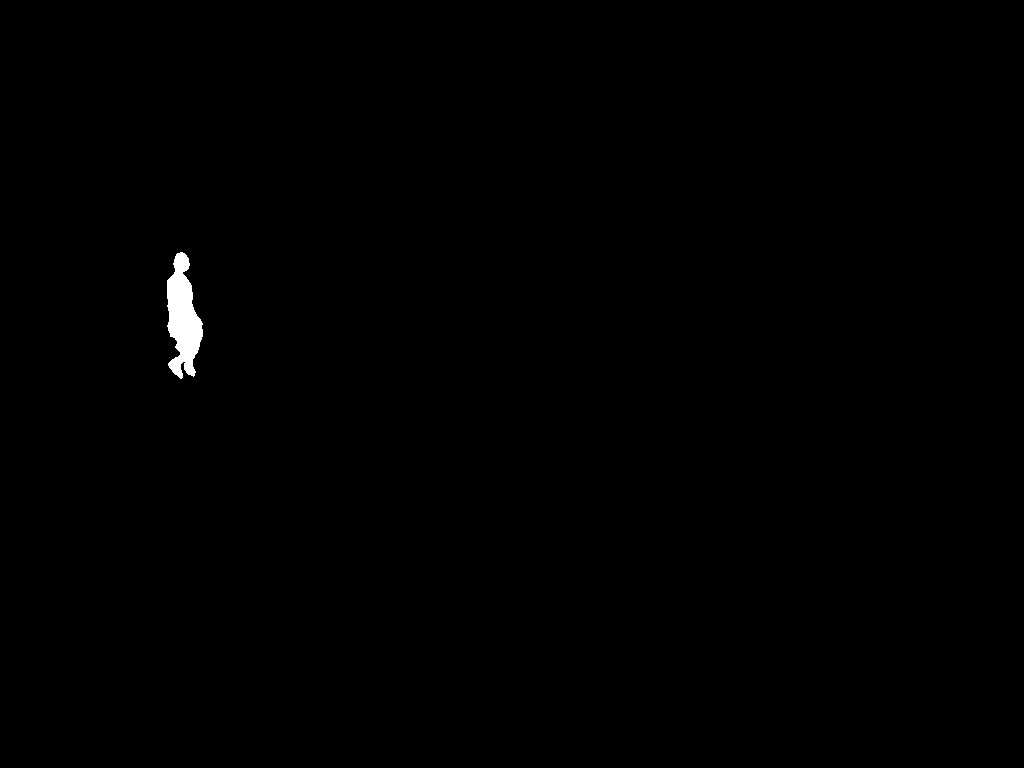} & 
\includegraphics[width=.16\textwidth, height=2.1cm]{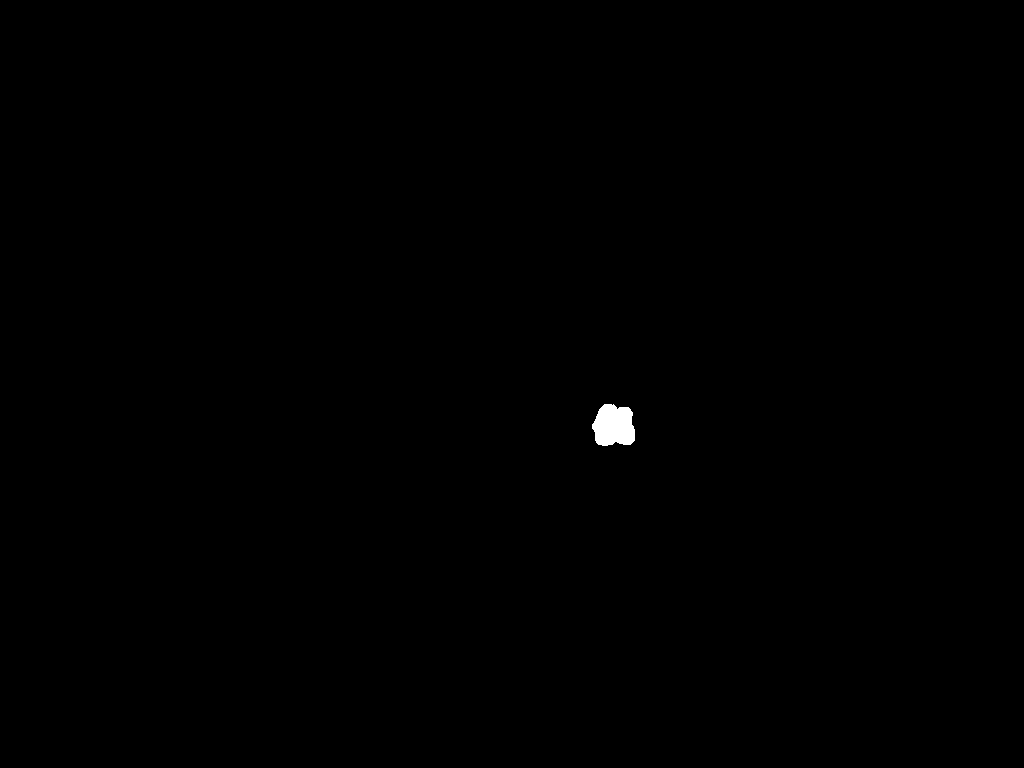} &  
\includegraphics[width=.16\textwidth, height=2.1cm]{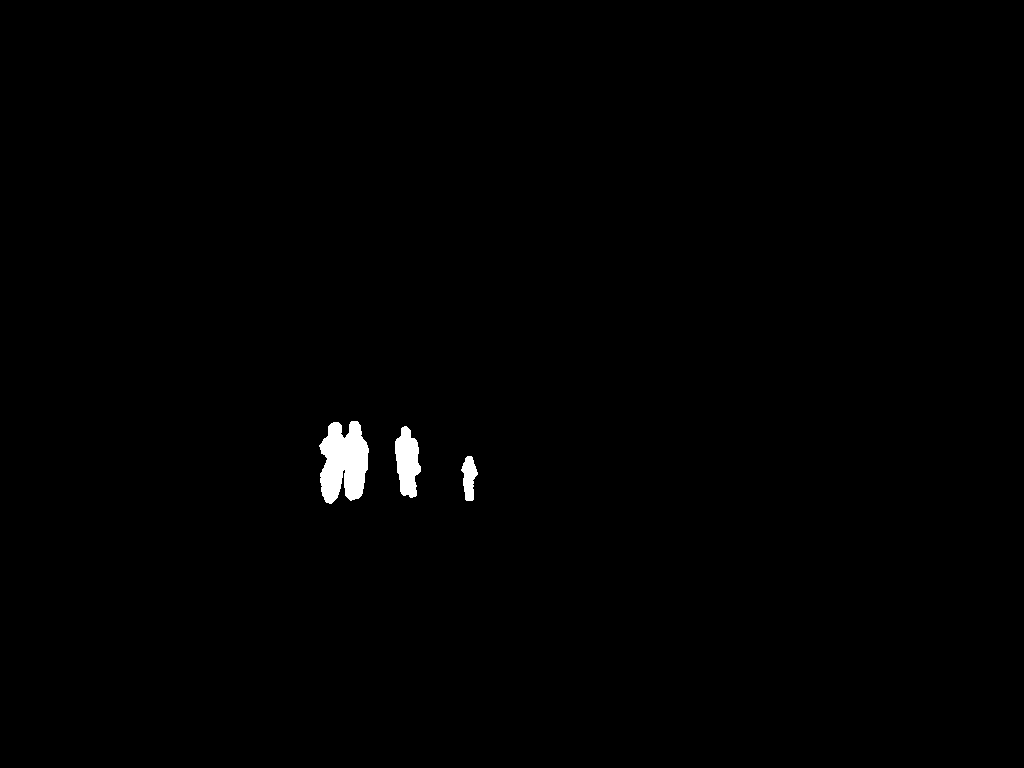} &  
\includegraphics[width=.16\textwidth, height=2.1cm]{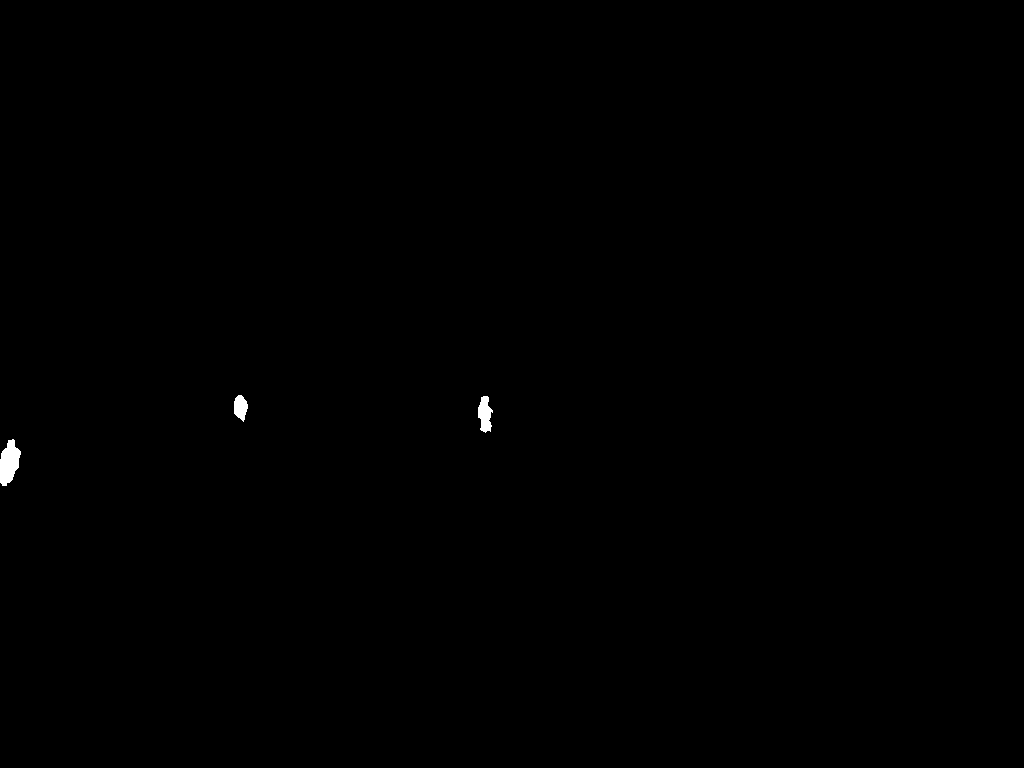} \\ 

\rotatebox{90}{\scriptsize{BASNet Th \cite{qin2019basnet}}} & \includegraphics[width=.16\textwidth, height=2.1cm]{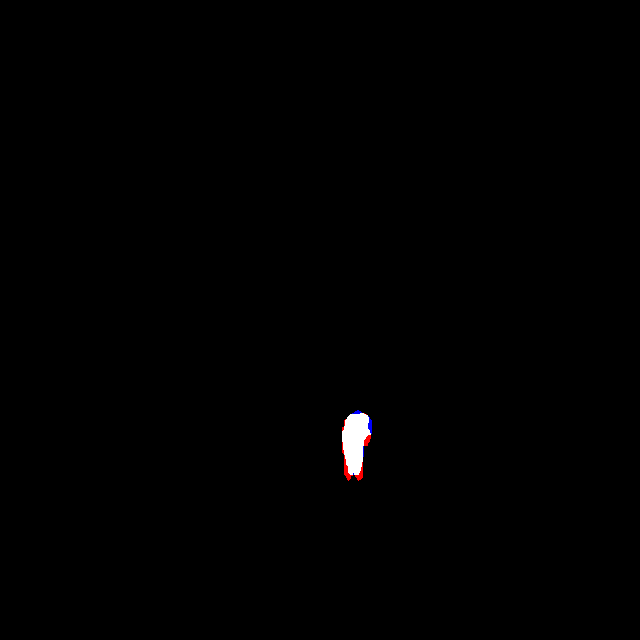} & 
\includegraphics[width=.16\textwidth, height=2.1cm]{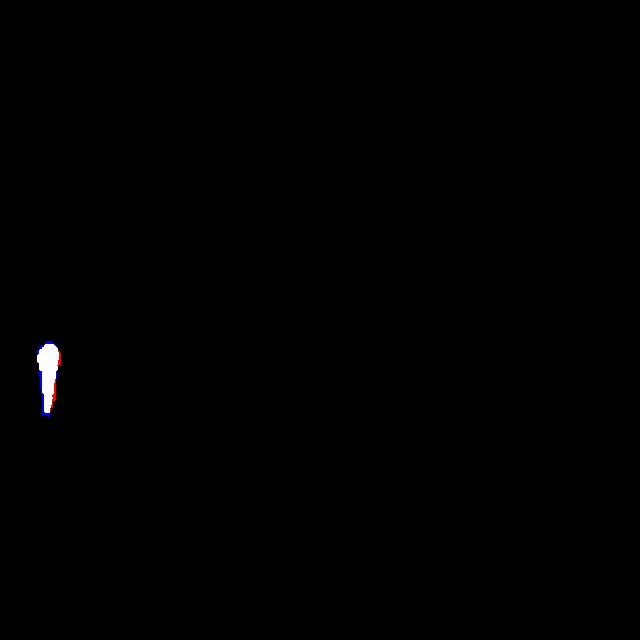} & 
\includegraphics[width=.16\textwidth, height=2.1cm]{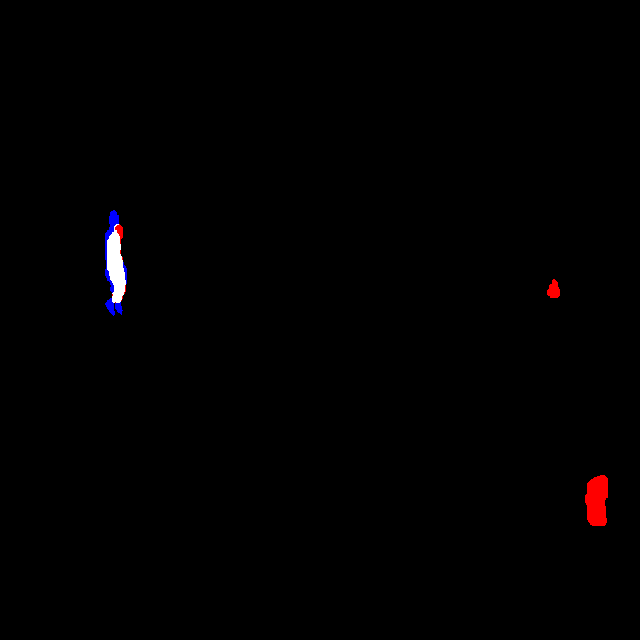} & 
\includegraphics[width=.16\textwidth, height=2.1cm]{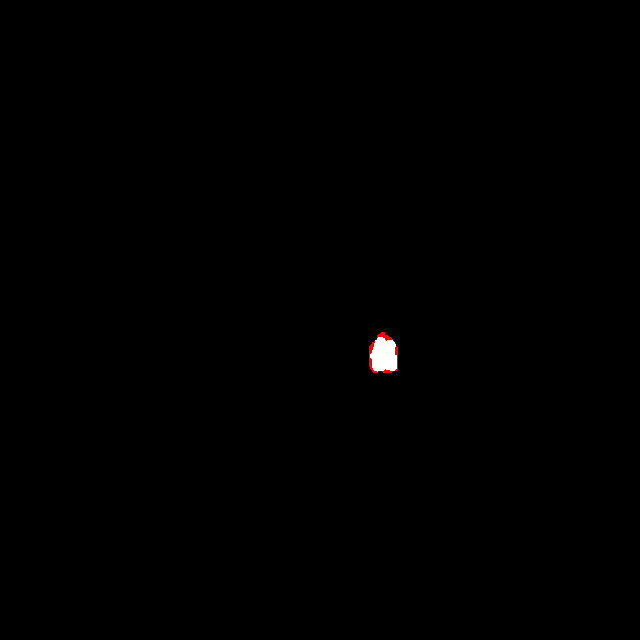} & 
\includegraphics[width=.16\textwidth, height=2.1cm]{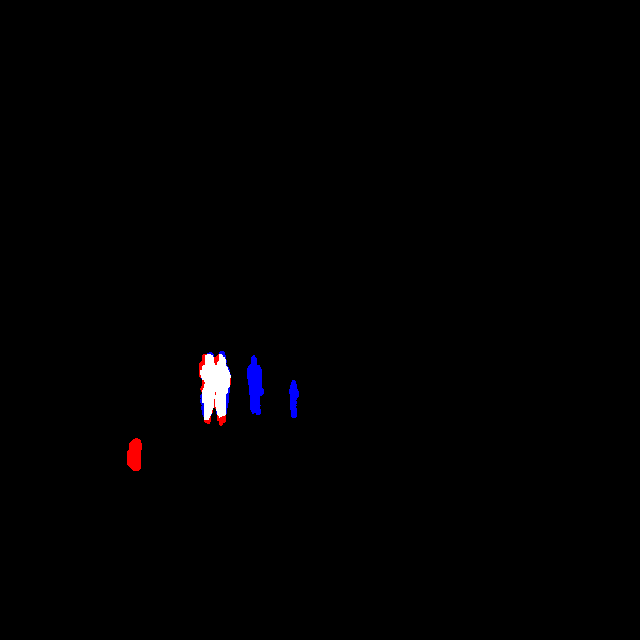} &  
\includegraphics[width=.16\textwidth, height=2.1cm]{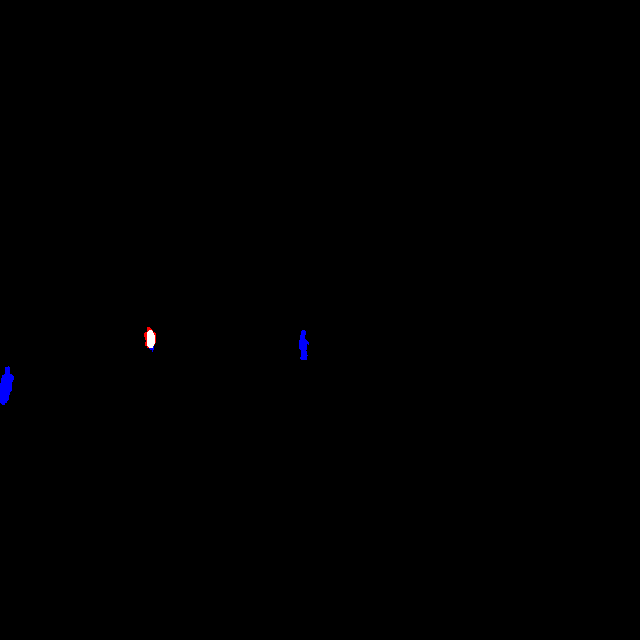} \\ 

\rotatebox{90}{\scriptsize{BGNet Th \cite{chen2022boundary}}} & \includegraphics[width=.16\textwidth, height=2.1cm]{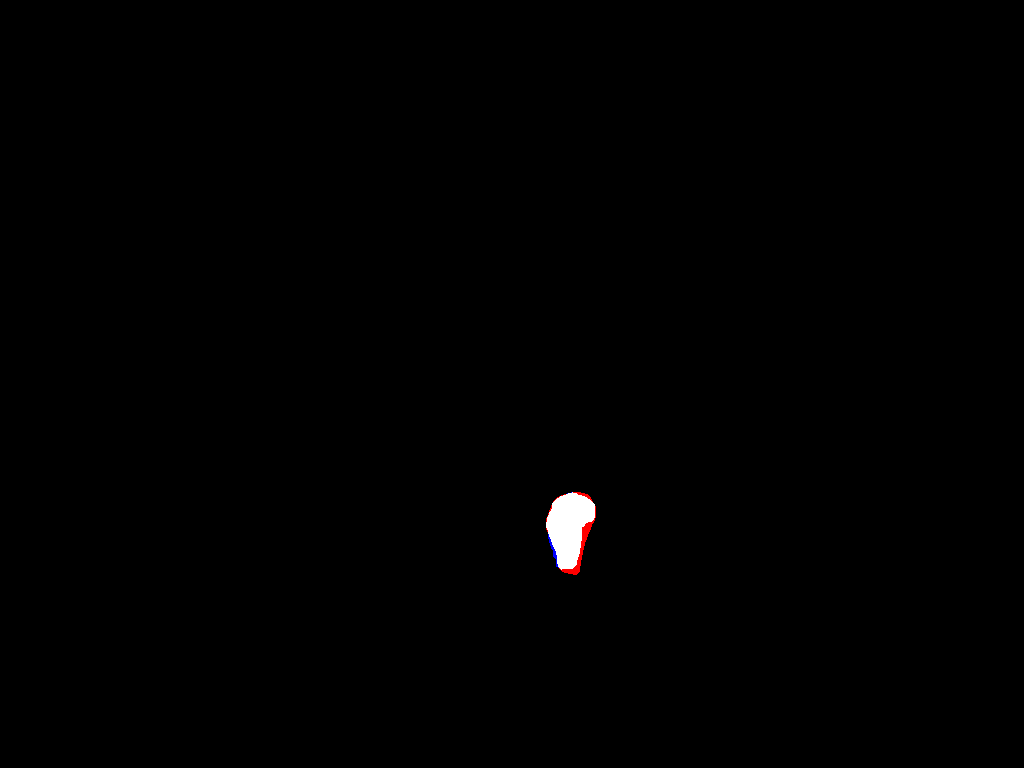} & 
\includegraphics[width=.16\textwidth, height=2.1cm]{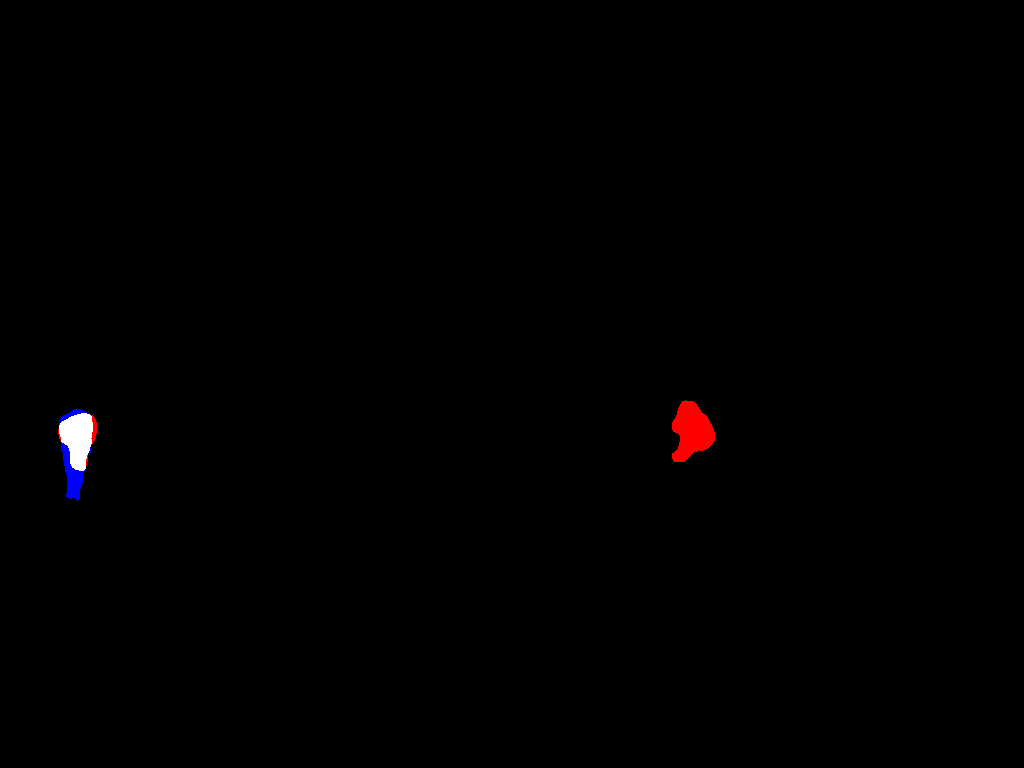} & 
\includegraphics[width=.16\textwidth, height=2.1cm]{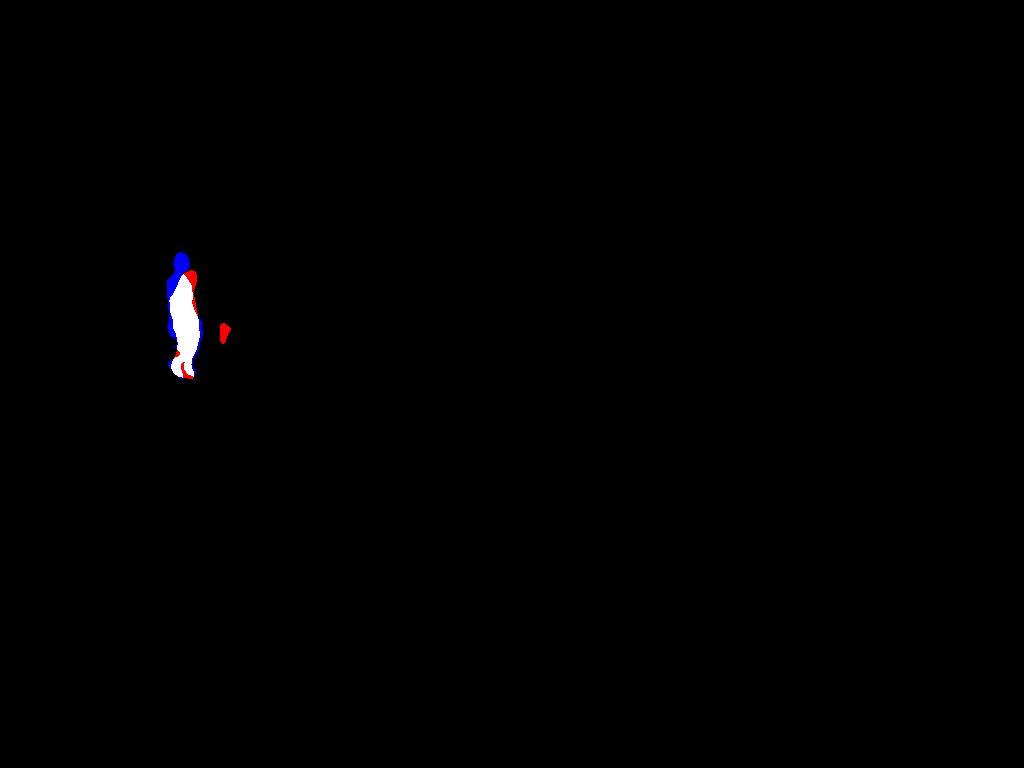} & 
\includegraphics[width=.16\textwidth, height=2.1cm]{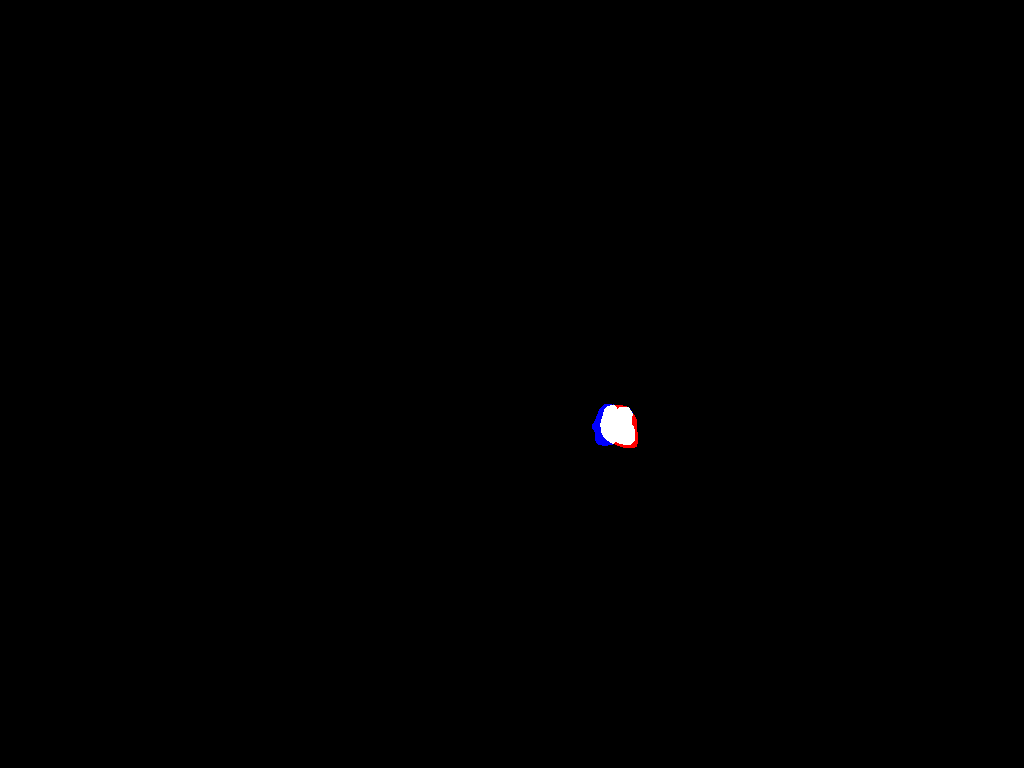} & 
\includegraphics[width=.16\textwidth, height=2.1cm]{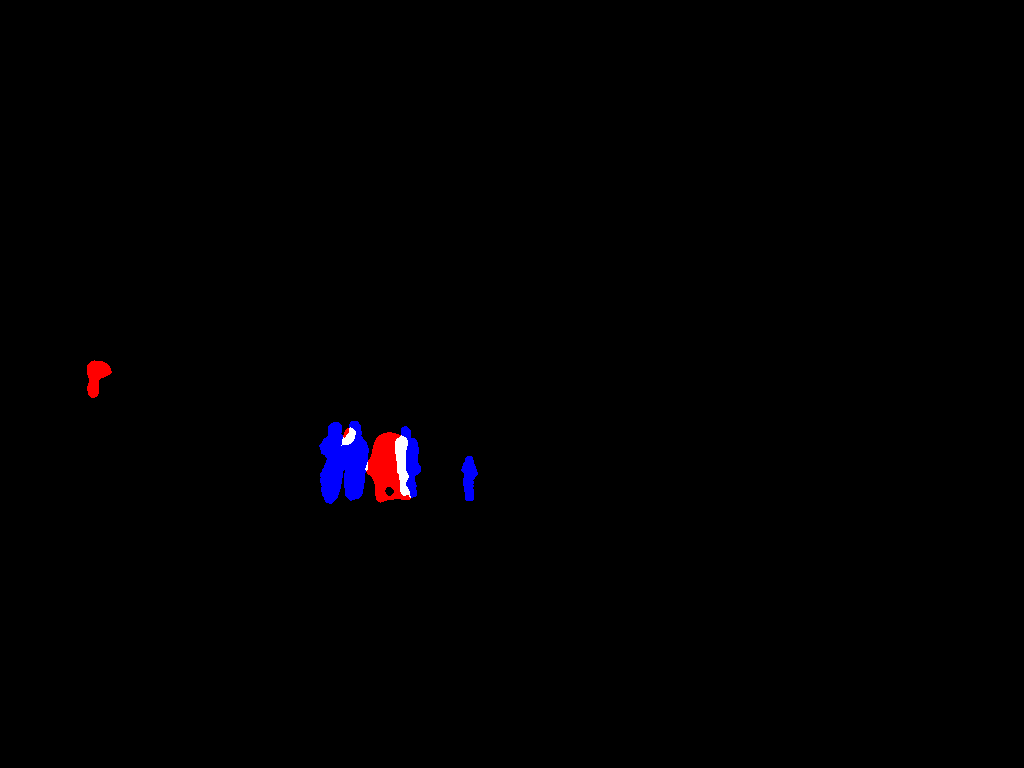} &  
\includegraphics[width=.16\textwidth, height=2.1cm]{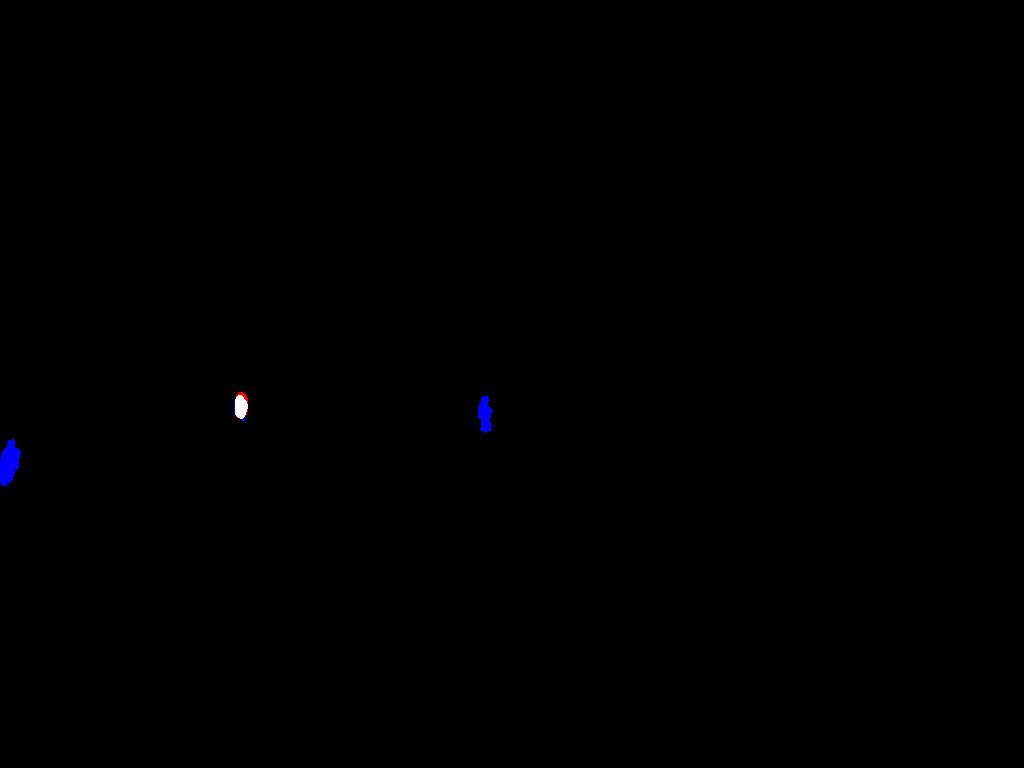} \\ 

\rotatebox{90}{\scriptsize{OCENet Th \cite{liu2022modeling}}} & \includegraphics[width=.16\textwidth, height=2.1cm]{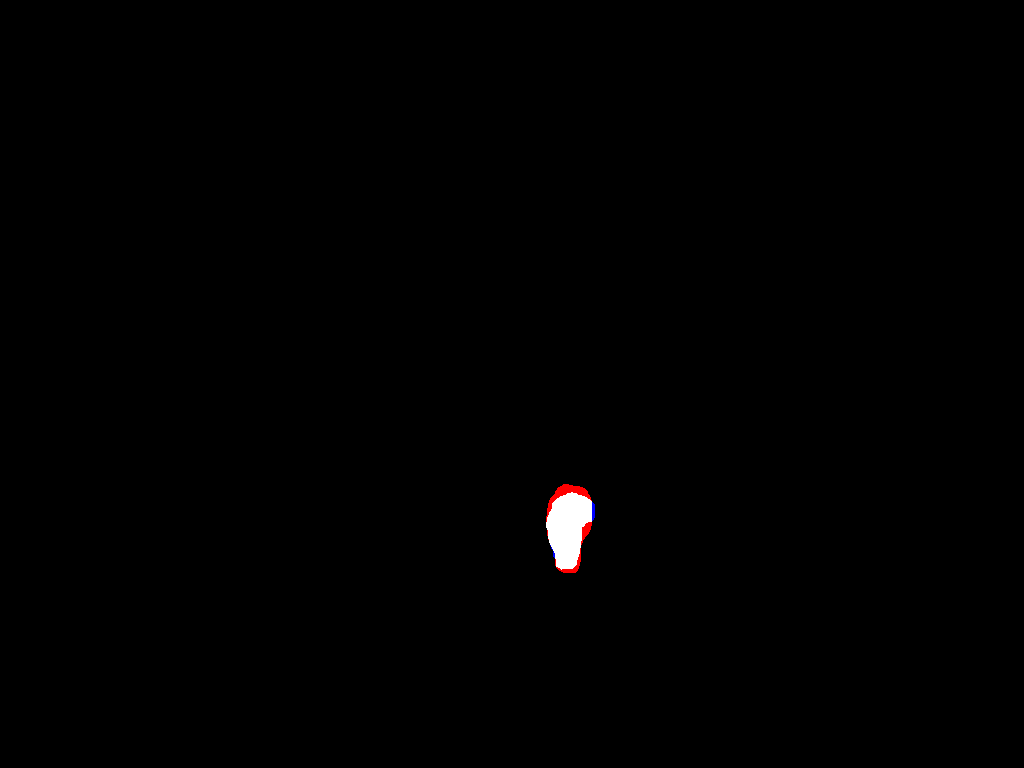} & 
\includegraphics[width=.16\textwidth, height=2.1cm]{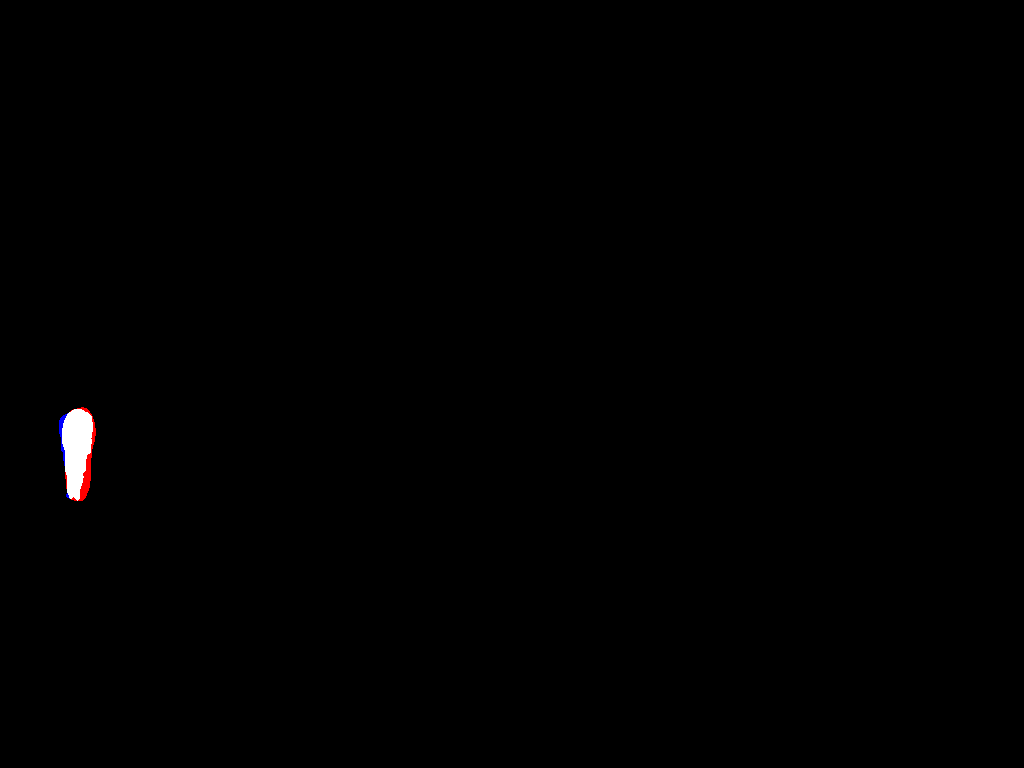} & 
\includegraphics[width=.16\textwidth, height=2.1cm]{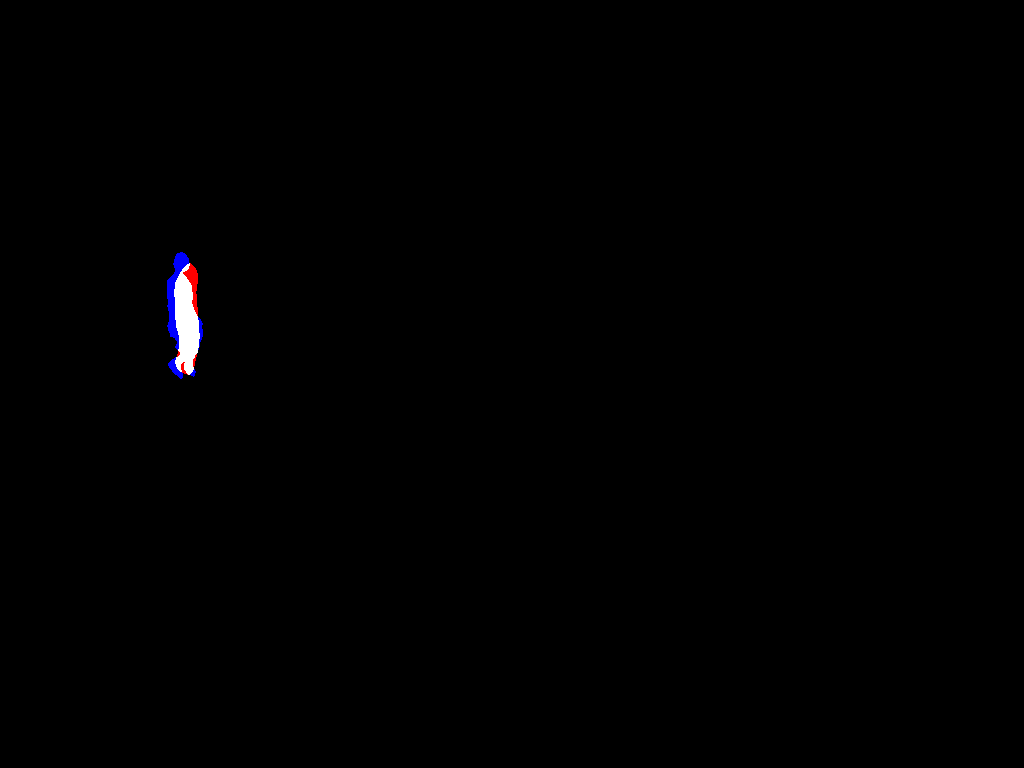} & 
\includegraphics[width=.16\textwidth, height=2.1cm]{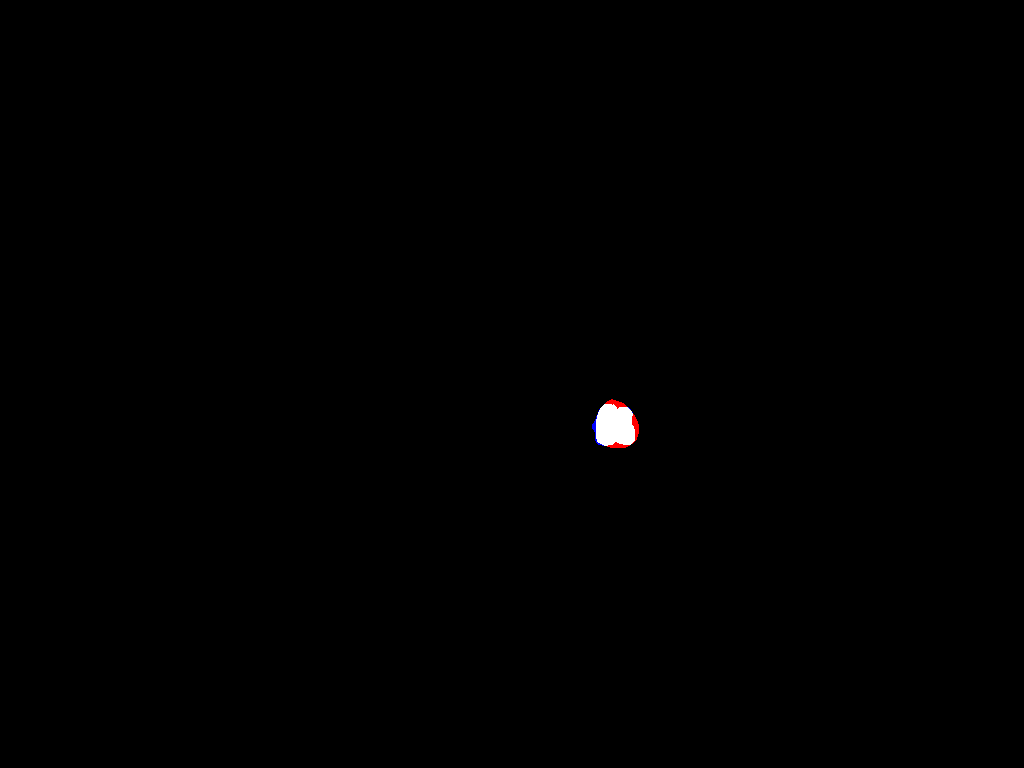} & 
\includegraphics[width=.16\textwidth, height=2.1cm]{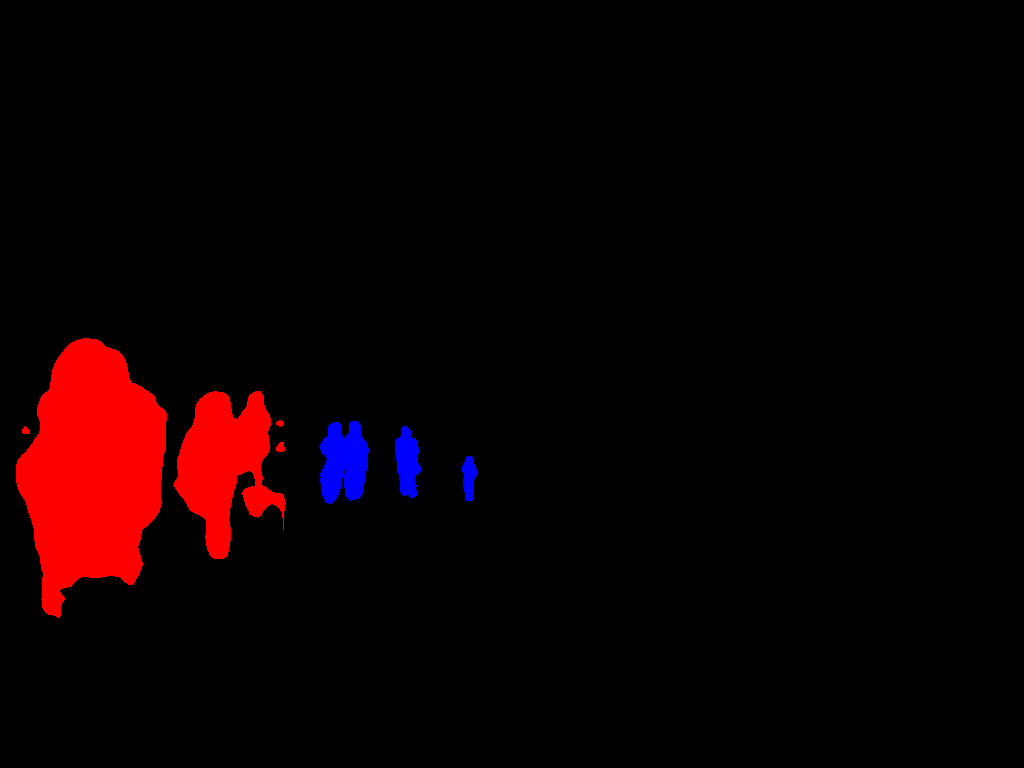} &  
\includegraphics[width=.16\textwidth, height=2.1cm]{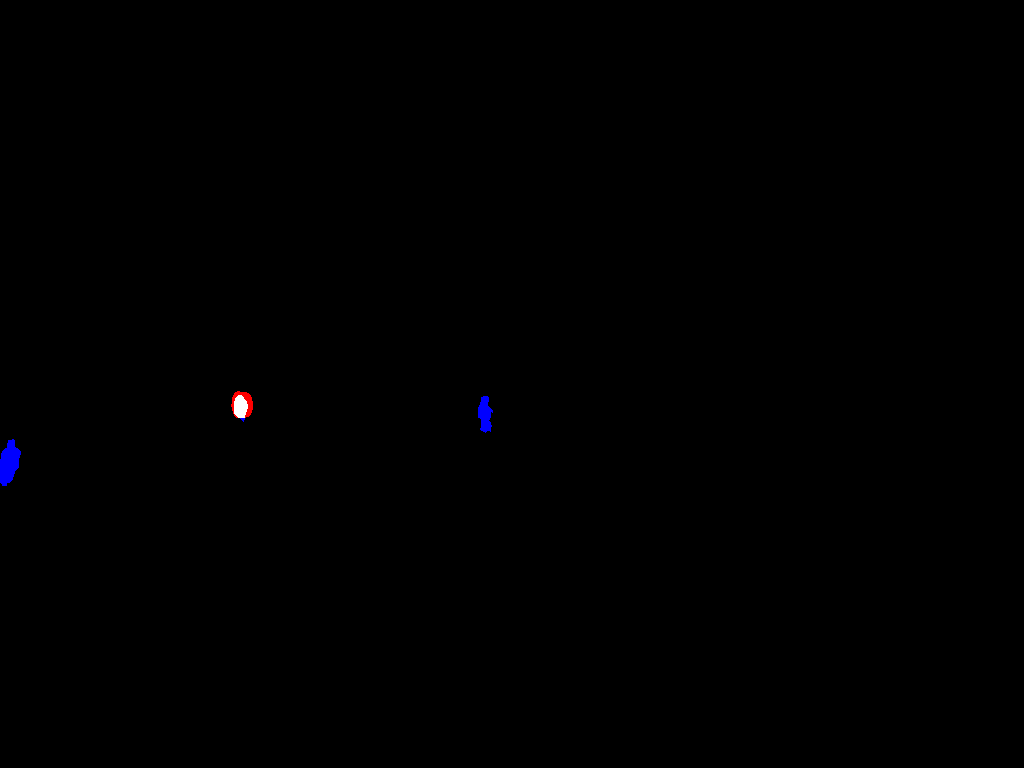} \\ 

\rotatebox{90}{\scriptsize{AVNet Vis \cite{velesaca2026iguana}}} & \includegraphics[width=.16\textwidth, height=2.1cm]{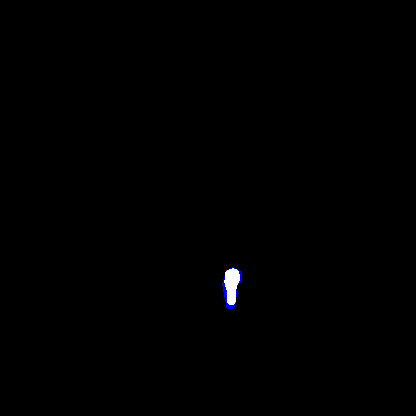} &
\includegraphics[width=.16\textwidth, height=2.1cm]{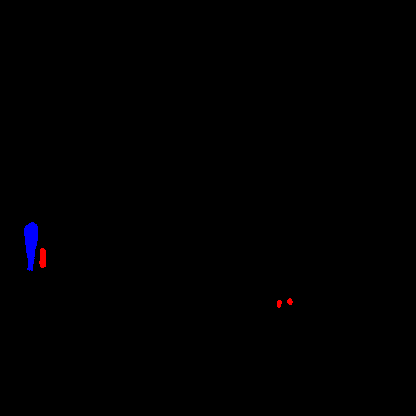} &
\includegraphics[width=.16\textwidth, height=2.1cm]{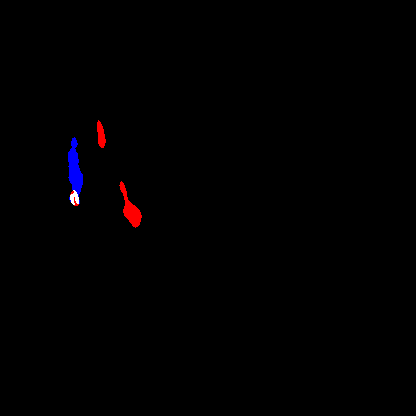} &
\includegraphics[width=.16\textwidth, height=2.1cm]{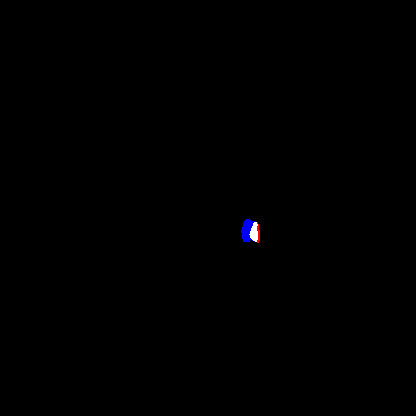} &
\includegraphics[width=.16\textwidth, height=2.1cm]{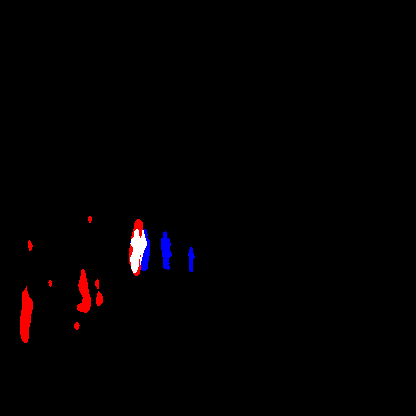} &
\includegraphics[width=.16\textwidth, height=2.1cm]{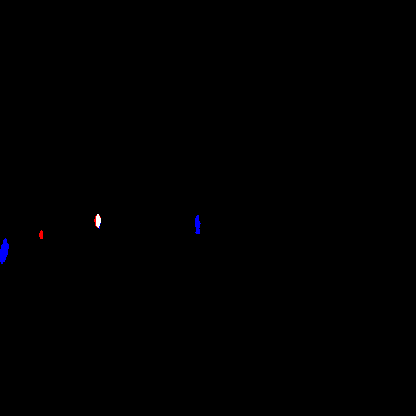} \\ 

\rotatebox{90}{\scriptsize{AVNet Th \cite{velesaca2026iguana}}} & \includegraphics[width=.16\textwidth, height=2.1cm]{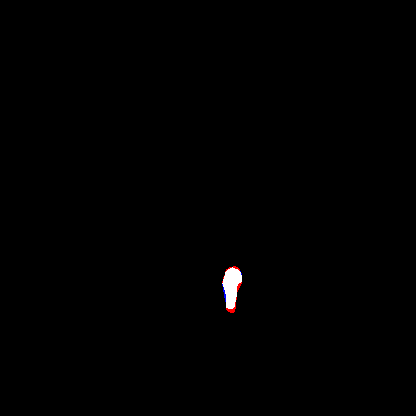} &
\includegraphics[width=.16\textwidth, height=2.1cm]{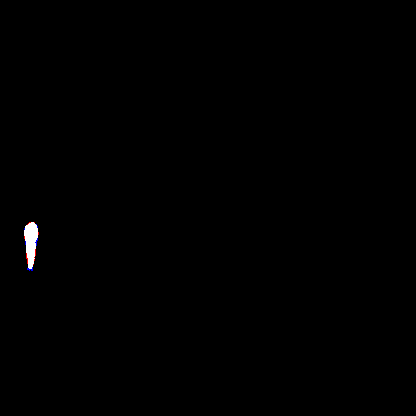} &
\includegraphics[width=.16\textwidth, height=2.1cm]{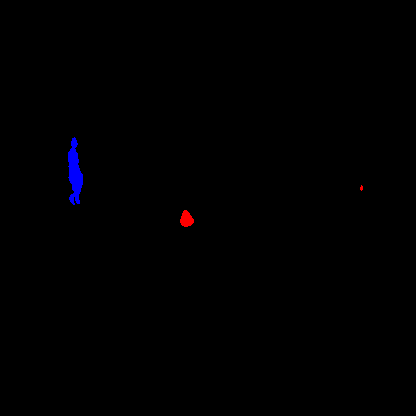} &
\includegraphics[width=.16\textwidth, height=2.1cm]{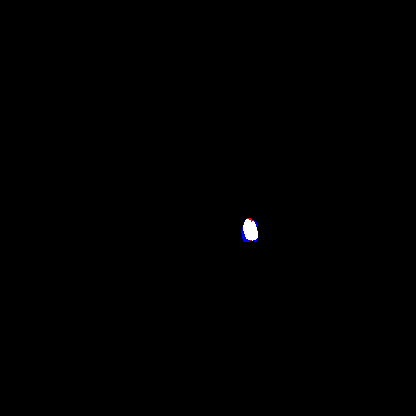} &
\includegraphics[width=.16\textwidth, height=2.1cm]{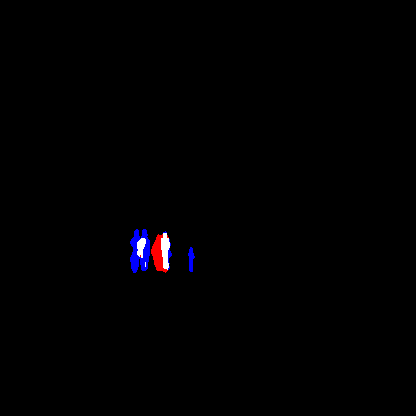} &
\includegraphics[width=.16\textwidth, height=2.1cm]{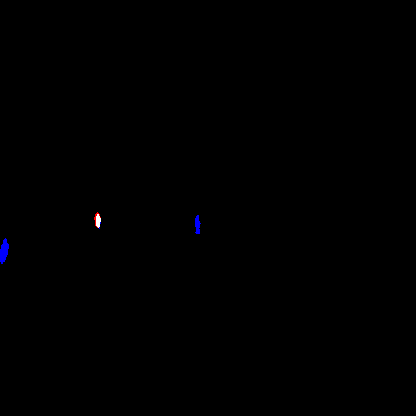} \\ 

\rotatebox{90}{\scriptsize{AVNet Vis+Th \cite{velesaca2026iguana}}} & \includegraphics[width=.16\textwidth, height=2.1cm]{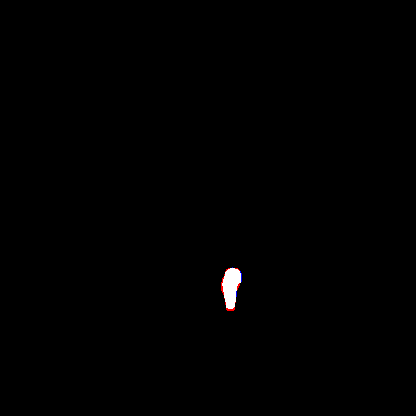} &
\includegraphics[width=.16\textwidth, height=2.1cm]{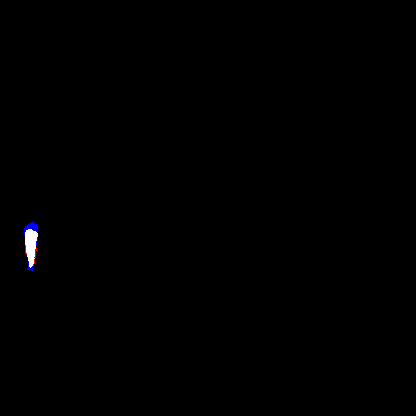} &
\includegraphics[width=.16\textwidth, height=2.1cm]{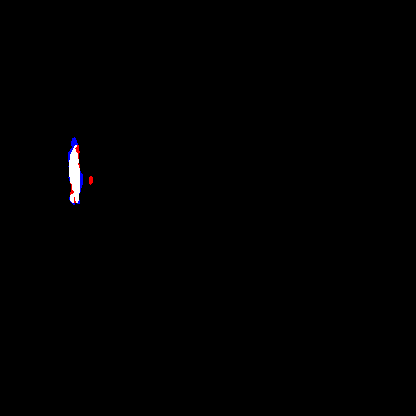} &
\includegraphics[width=.16\textwidth, height=2.1cm]{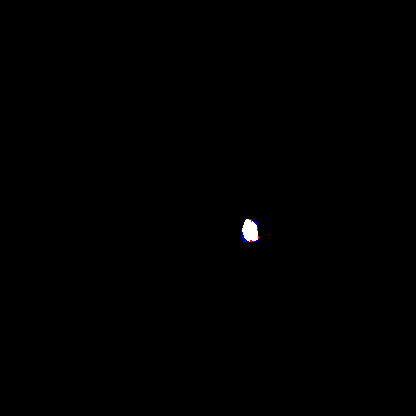} &
\includegraphics[width=.16\textwidth, height=2.1cm]{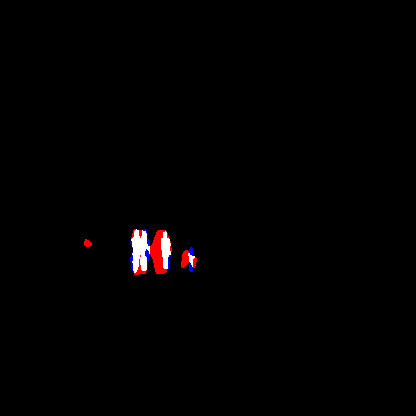} &
\includegraphics[width=.16\textwidth, height=2.1cm]{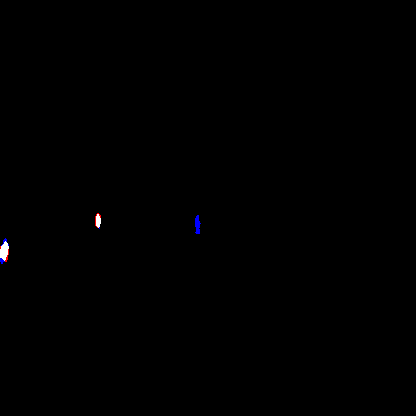} \\ 

\end{tabular}
}
\caption{Results using SoTA COD techniques that have achieved first or second place in at least one of the metrics. Successful matches between GT and predicted masks (white areas); False positive regions (red areas, over-segmentation); and false negative regions (blue areas, miss-segmentation).}
\label{fig:results}
\end{figure*}

\section{Experimental Results}
\label{sec:exp}
The experimental evaluation of the Camo-M3FD dataset utilizes a comprehensive suite of SoTA COD models to establish a robust performance baseline. This section analyzes the quantitative metrics across different modalities and provides a qualitative assessment of the segmentation challenges inherent in cross-spectral pedestrian camouflage.

\subsection{Quantitative Evaluation}
As summarized in Table \ref{tab:results_cod_camo-m3fd}, the quantitative results underscore the significant impact of spectral modality on detection accuracy. The quantitative evaluation across various state-of-the-art architectures reveals several critical trends regarding the performance of camouflaged object detection on the Camo-M3FD dataset. A consistent observation across all evaluated models is that the thermal modality significantly outperforms the visible baseline, suggesting that thermal signatures provide indispensable cues for localizing pedestrians when their visual appearance blends seamlessly into the background. Among the evaluated techniques, AVNet consistently achieves the highest performance, particularly when leveraging its native multispectral integration. It achieves the highest scores in nearly all key metrics, including a $S_{\alpha}$ of 0.7318, a weighted F-measure ($F_{\beta}^{w}$) of 0.5301, and an enhanced-alignment measure ($E^{mean}_{\phi}$) of 0.8287. This underscores the effectiveness of cross-spectral fusion in resolving complex camouflage scenarios that a single modality cannot fully address.

Furthermore, models that incorporate explicit boundary modeling or uncertainty awareness, such as OCENet and BGNet, demonstrate high competitive resilience in the thermal domain, exhibiting strong structural alignment and precision. The error margins across the top-tier models remain notably low, indicating high spatial accuracy in the predicted masks even under challenging environmental blending. Overall, these results confirm that while single-modality models show a clear preference for thermal data, the most robust detection is achieved by effectively utilizing the complementary nature of both spectra.

\subsection{Qualitative Evaluation} 
The qualitative analysis presented in Figure \ref{fig:results} reinforces the metric-based findings by visualizing the segmentation successes and failures of the leading models. A primary observation is the prevalence of "miss-segmentation" (blue areas) in visible-only or thermal-only predictions, where models fail to detect extremities or entire torso sections that blend seamlessly with the background texture. In contrast, predictions from thermal-based models and the multispectral AVNet exhibit much higher structural coherence, successfully capturing the human silhouette. However, the dataset presents a persistent challenge in the form of "over-segmentation" (red areas), where models erroneously include background elements with similar heat signatures or edge profiles as part of the pedestrian mask. The results from AVNet (Vis+Th) show the most accurate alignment with the ground truth (white areas), effectively using the visible channel to refine boundaries that might be blurred in the thermal domain. These results collectively validate Camo-M3FD as a challenging benchmark that requires sophisticated multimodal fusion to achieve high-precision pedestrian detection in complex environments.

\section{Discussion}
\label{sec:disc}
The results obtained from the proposed Camo-M3FD benchmark provide critical insights into the nature of pedestrian camouflage across different spectral domains. A primary finding is the inherent limitation of visible-spectrum (RGB) sensors when dealing with advanced camouflage, where the foreground-background similarity in texture and color often leads to significant miss-segmentation. Our experiments demonstrate that the integration of thermal imaging is not merely an enhancement but a necessity for reliable detection in these scenarios. The thermal modality effectively "breaks" the visual camouflage by highlighting heat signatures that are nearly impossible to mask with traditional visual techniques.

However, the "over-segmentation" observed in several high-performing models—where non-pedestrian heat sources are incorrectly classified—indicates that thermal data alone is not a definitive solution. This suggests that future research should focus on more sophisticated cross-modal interaction modules. Specifically, architectures that can dynamically weigh the importance of each spectrum based on environmental conditions (e.g., thermal noise in hot weather vs. visual clutter) are likely to define the next generation of COD models. Furthermore, the performance gap between existing SoTA models on our dataset compared to traditional COD benchmarks confirms that Camo-M3FD presents a more complex and realistic challenge, moving beyond static biological camouflage into dynamic, real-world pedestrian scenarios.

\section{Conclusions}
\label{sec:conclu}
This paper introduces Camo-M3FD, a novel benchmark dataset specifically designed for cross-spectral camouflaged pedestrian detection. An extensive evaluation of state-of-the-art models establishes a clear baseline for future work, highlighting the superior performance of multispectral fusion approaches, such as AVNet, over single-modality methods. The findings emphasize that while thermal signals provide indispensable cues for localization, the visual domain remains crucial for refining structural details. It is expected that Camo-M3FD will serve as a foundational resource for the community, encouraging the development of more robust, safety-critical detection systems for autonomous driving and surveillance.

\section*{Acknowledgements}
This material is based upon work supported by the Air Force Office of Scientific Research under award number FA9550-24-1-0206; and partially supported by the Grant PID2021-128945NB-I00 funded by MICIU/AEI/ 10.13039/501100011033 and by ERDF/EU and Grant PID2024-162815NB-I00 funded by MICIU/AEI/ 10.13039/501100011033 and by ERDF/EU; and by the ESPOL project ``Advancing Camouflaged Object Detection with a cost-effective Cross-Spectral vision system (ACODCS)'' (CIDIS-003-2024). The authors acknowledge the support of the Generalitat de Catalunya CERCA Program to CVC’s general activities.

{
    \small
    \bibliographystyle{ieeenat_fullname}
    \bibliography{main}
}


\end{document}